\documentclass{article}

\usepackage[preprint]{neurips_2025}

\usepackage[utf8]{inputenc} 
\usepackage[T1]{fontenc}    
\usepackage{hyperref}       
\usepackage{url}            
\usepackage{booktabs}       
\usepackage{amsfonts}       
\usepackage{nicefrac}       
\usepackage{microtype}      
\usepackage{xcolor}         
\usepackage{amsmath} 
\usepackage{graphicx} 
\usepackage[most]{tcolorbox}

\usepackage{algorithm}
\usepackage{algpseudocode}
\usepackage{float}

\title{Steering Conceptual Bias via Transformer Latent-Subspace Activation}

\author{%
  Vansh Sharma\thanks{Corresponding Author: vanshs@umich.edu} \\
  University of Michigan\\
  \And
  Venkat Raman \\
  University of Michigan\\
}

\begin{document}

\maketitle

\begin{figure}[!htb]
  \centering
  \includegraphics[width=1.0\linewidth]{ 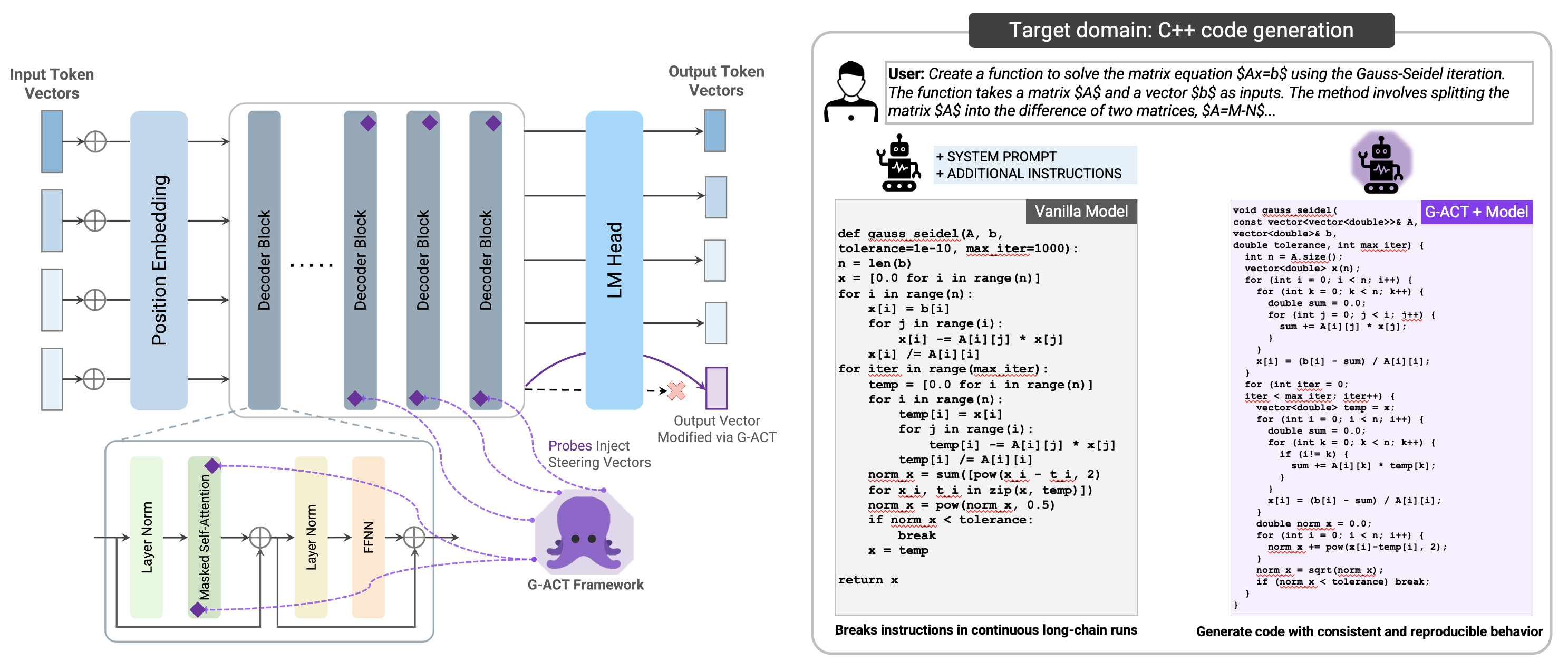}
\end{figure}

\begin{abstract}
    This work examines whether activating latent subspaces in language models (LLMs) can steer scientific code generation toward a specific programming language. Five causal LLMs were first evaluated on scientific coding prompts to quantify their baseline bias among four programming languages. A static neuron-attribution method, perturbing the highest activated MLP weight for a `C++ or CPP' token, proved brittle and exhibited limited generalization across prompt styles and model scales. To address these limitations, a gradient-refined adaptive activation steering framework (G-ACT) was developed: per-prompt activation differences are clustered into a small set of steering directions, and lightweight per-layer probes are trained and refined online to select the appropriate steering vector. In LLaMA-3.2 3B, this approach reliably biases generation towards the CPP language by increasing the average probe classification accuracy by 15\% and the early layers (0–6) improving the probe classification accuracy by 61.5 \% compared to the standard ACT framework. For LLaMA-3.3 70B, where attention-head signals become more diffuse, targeted injections at key layers still improve language selection. Although per-layer probing introduces a modest inference overhead, it remains practical by steering only a subset of layers and enables reproducible model behavior. These results demonstrate a scalable, interpretable and efficient mechanism for concept-level control for practical agentic systems.

\end{abstract}

Neuron Attribution, Privileged Basis, Code Generation, Reproducibility

\section{Introduction}
Large language models (LLMs) have rapidly evolved into sophisticated natural language processors, enabling the development of agentic systems that autonomously orchestrate complex workflows \citep{wang2024survey, vanshRAG}. A particularly striking trend is the adoption of LLM‑driven agents for automated code generation. By incorporating plugin architectures that expose external APIs, these agents extend beyond text synthesis: agents can invoke specialized tools \citep{wang2024executableCoadeAct} and execute command-line operations \citep{openAI_codex,lu2024aiscientistfullyautomatedSakana}. This expanded action space has already powered a variety of real-world applications - from LLM-based robotic controllers \citep{wang2023voyageropenendedembodiedagent, KannanPurdue, ZiyiBrownLLMRobots} to automated scientific experimentation \citep{gao2024empowering, gridach2025agentic} platforms - highlighting the remarkable success of these agents. 

Despite the growing utility of LLM agents for high‑level scripting and automation, their application to scientific code generation remains unexplored. Scientific software predominantly relies on C++ or CPP, CUDA and other low‑level languages that are sparsely represented in most pretraining datasets. As a result, LLM‑generated implementations often exhibit syntactic or semantic errors, leading to compilation failures or unstable behavior at runtime \citep{zhang2023planning}. In addition, current agents are heavily dependent on user-defined control primitives and meticulously engineered prompts, which can be misinterpreted and give rise to unpredictable execution paths \citep{kim2025promptflowintegrityprevent, krishnan2025aiagentsevolutionarchitecture, gridach2025agentic}. One possible solution is to augment function definitions and syntax using a retrieval augmented generation (RAG) framework; however, \citet{xiao2023efficient} and \citet{yona2025interpretingDeepMind} report unexplained vulnerability linked to "attention sinks" or first token bias, causing model behavior to diverge due to token repetition in long interactions, commonly seen in RAG. Another failure mode arises from the use of LMs that have been trained or fine‑tuned on undisclosed corpora and then subjected to opaque alignment procedures, especially for MoE models \citep{nishu2025denseApple}. Such processes can inadvertently skew the model's output toward a particular programming language or coding style, further eroding its ability to generate correct generalizable code across the various low-level languages prevalent in scientific computing during agentic applications with long repeated interactions. 

To investigate these bottlenecks, a curated benchmark of scientific coding challenges is introduced to reveal the implicit language preference of an LLM when presented with a given problem. Targeted probing techniques are then applied to identify subgraphs or subspaces in the model whose activation strongly correlates with that preference. 
In the \emph{privileged} basis of the model \citep{zhang2024towardsActivation}, often termed native MLP activation axes, each coordinate encodes a distinct functional characteristic (here the coding language preference). This basis is a consequence of elementwise non-nonlinearities (e.g., ReLU, GELU) breaking rotational symmetry and the coordinate‑wise biases introduced by optimizers such as Adam \citep{anthropicBasisExistence}. Since these axes remain (approximately) disentangled, selectively amplifying or suppressing a single coordinate produces clear causal shifts in token probabilities \citep{anthropicBasisToy}. In this basis, an effective weight vector for any neuron can be derived and its direct influence on features can be decoded. The external perturbation of this neuron’s activation along the identified axis then reliably steers the model toward generating code in the desired programming language. 

Contribution of this work:
\begin{enumerate}
    \item A curated suite of simple and complex scientific programming prompts designed to systematically evaluate programming language selection behavior across targeted coding tasks.

    \item A proof‐of-concept method that locates the single MLP weight most correlated with the target concept (CPP language in this case) and perturbs it to bias model preference, demonstrating the feasibility and fragility of manual neuron edits for style control.

    \item A scalable method that trains lightweight per-layer probes refined during inference via cross-entropy loss. This approach reliably steers LLMs toward the CPP language (or any target domain) with minimal runtime overhead and embeds reproducible transformation matrices for consistent behavior across multiple model deployments.
\end{enumerate}

\section{Related Work} \label{relatedW}
Reverse engineering neural networks to drive/steer a specific behavior, known as mechanistic interpretability, is an emerging field \citep{wang2022interpretability, templeton2023_claude, olsson2022context}. Its primary objective is to identify causal relationships within model activations \cite{jiang2024interpreting, chen2024robust, rodrigues2024unsupervised} thereby revealing complex functional roles \citep{condori2021analysis, szandala2023unlocking} and enabling the targeted modulation \citep{huang2025pcg} or ablation of individual neurons. Supervised fine-tuning, weight modulation techniques, and RLHF represent direct intervention strategies for model steering \citep{dathathri2019plug, meng2022locating}. Although effective, these methods impose substantial computational overhead and can inadvertently compromise the robustness and general performance of the model \citep{brown2023robustness}.

The method using corrupted inputs as a baseline distribution to resample neuron activation, known as Activation Patching, has been widely used to achieve fine-grained control over the model output \citep{kramar2024atp, geiger2022inducing, meng2022locating}. A typical causal attribution method quantifies the negative impact on model output by deleting specific neuron activations, a more trivial approach compared to Activation Patching. In such methods, extensive model sweeps are conducted to evaluate the results of neuron modification, leading to millions of model evaluations \citep{kramar2024atp}. Studies such as \citep{park2025steer, yeo2025debiasing}, have focused on suppressing hallucinations in LLMs within different modalities using activation patching. Recent studies using similar forms of neuron attribution have typically been applied to multiple choice question benchmarks \citep{act, davies2025flashinterpretability}, rather than real-world deployment scenarios. In contrast to previous work, this study presents a different approach for the inference-time adaptive steering algorithm \citep{act} by incorporating gradient-based refinement, offering a more efficient and precise mechanism for steering large-scale models.

\section{Methodology}
\label{method}
Autoregressive decoder-only transformers convert embedded tokens into high‑dimensional latent representations that are successively refined by the attention mechanism \citep{Vaswaniattentionisall} and deep MLP blocks for output generation. Residual connections carry a persistent residual stream: combining each MLP input with its output to preserve and enhance contextual features before the LM head decodes the final latent stream into tokens. In particular, individual neurons operate directly on the residual stream, a central information pathway through transformer layers. Neurons specialized in the detection of specific features within this stream exhibit a strong alignment between their weight vectors and the corresponding features, characterized by high cosine similarity \citep{meng2022locating}. This property forms the basis for the methods explained in the following sections.

\subsection{Neuron Attribution: Static Method}\label{sec:mlp_static}
Building upon the previously mentioned alignment property between weights and features, the static method involves directly decoding the neuron weight vectors through the LM-head, thus converting the neuron weights into interpretable token-level probability distributions. This approach aligns conceptually with the ``Golden Gate" investigation by \cite{templeton2023_claude} and work by \cite{davies2025flashinterpretability}.

A transformer with $L$ layers processes an input token sequence $x_{1},\dots,x_{n}\in V$ (vocabulary $V$) through a residual-stream architecture. Each token $x_{i}$ is first embedded into a $d$-dimensional vector $h_{i}^{(0)}$. In layer $j\in\{1,\ldots,L\}$, the residual update at position $i$ is given by
\[
h_{i}^{(j)} \;=\; h_{i}^{(j-1)} \;+\; f^{(j)}\bigl(h_{1}^{(j-1)},\dots,h_{n}^{(j-1)}\bigr)\,,
\]
where $f^{(j)}$ denotes the combined self-attention and feedforward operations of layer $j$.  After $L$ such updates, the final residuals $h_{i}^{(L)}$ are passed through a learned linear head (LM-head or "output head" or "prediction head") and softmax to produce next token probabilities:
\[
P\bigl(x_{i+1}\mid x_{1:i}\bigr)
\;=\;\mathrm{softmax}\bigl(W\,h_{i}^{(L)}\bigr)\,.
\]
To identify neurons selectively responsive to a particular target token, the decoded probability corresponding to that token is normalized by the average decoded probability of the top $k$ tokens for each neuron. Upon identifying candidate feature-selective neurons, their functional significance is validated by systematically modifying their activations \citep{activationPatching}. For pre-trained models, the role of individual neurons can be probed through precisely crafted prompts that elicit the desired features, followed by controlled variations in neuron outputs to measure corresponding changes in token probabilities.  The neuron with the highest activation is identified and then its activation value is increased by a fixed amount, while all other coordinates remain unchanged. The modified residual vector is passed unchanged through the remaining layers of the transformer. The subsequent steps provide a detailed description of this approach. Here, the residual stream dimension (final hidden state) is $D$ and the vocabulary dimension is denoted as $V$.
\begin{enumerate}
    \item \textbf{Extraction from Transformer Layers}
    \begin{itemize}
        \item Iterate over each transformer layer, accessing the corresponding MLP components.
    \end{itemize}

    \item \textbf{Computation of Effective Neuron Weights}
    \begin{itemize}
        \item Calculate effective neuron weights using an elementwise multiplication of up-projection weights $\mathbf{W}_{up}$ and the sigmoid activation ($\sigma$) of gate-projection weights $\mathbf{W}_{gate}$:
        \begin{equation}
            \mathbf{W}_{\text{eff}} = \mathbf{W}_{up} \odot \sigma(\mathbf{W}_{gate})
        \end{equation}
    \end{itemize}

    \item \textbf{Decoding via LM Head}
    \begin{itemize}
        \item Decode each row ($\mathbf{w}_{\text{eff}}$) of $\mathbf{W}_{\text{eff}}$ using the LM head ($\mathbf{W}_{LM} \in \mathbb{R}^{V \times D}$), mapping it into vocabulary logits:
        \begin{equation}
            \text{Logits} = \mathbf{W}_{LM}\mathbf{w}_{\text{eff}} + \mathbf{b}_{LM}
        \end{equation}
        \item Compute probability distribution over the vocabulary using the softmax function:
        \begin{equation}
            P = \text{softmax}(\text{Logits})
        \end{equation}
    \end{itemize}

    \item \textbf{Normalized Activation Score Calculation}
    \begin{itemize}
        \item Determine probability $P_t$ for the token of interest $t$:
        \begin{equation}
            P_t = P[t]
        \end{equation}
        \item Calculate average probability $P_{avg}$ over the top-$k$ (here, $k=100$) tokens:
        \begin{equation}
            P_{avg} = \frac{1}{k}\sum_{i=1}^{k} P(i)
        \end{equation}
        \item Obtain the normalized activation score ($A_N$):
        \begin{equation}
            A_N = \frac{P_t}{P_{avg}}
        \end{equation}
    \end{itemize}
\end{enumerate}

\subsection{Neuron Attribution: Adaptive Method}\label{sec:mlp_adaptive}
The algorithm presented above targets individual neuron activations directly, but in large‐scale models the notion that a single neuron uniquely encodes a high‑level concept is often invalid: neurons are commonly \textit{polysemantic}, responding to many different features or tokens. To achieve consistent concept steering, one must, therefore, identify all neurons associated with a given concept - across every layer - and apply coordinated adjustments. Building on the Adaptive Activation Steering (ACT) framework \citep{act}, an adaptive, multi‑neuron gradient-based algorithm is proposed that, at inference time, activates the entire set of concept‑linked neurons.

\begin{figure}[!htb]
  \centering
  \includegraphics[width=.80\linewidth]{ 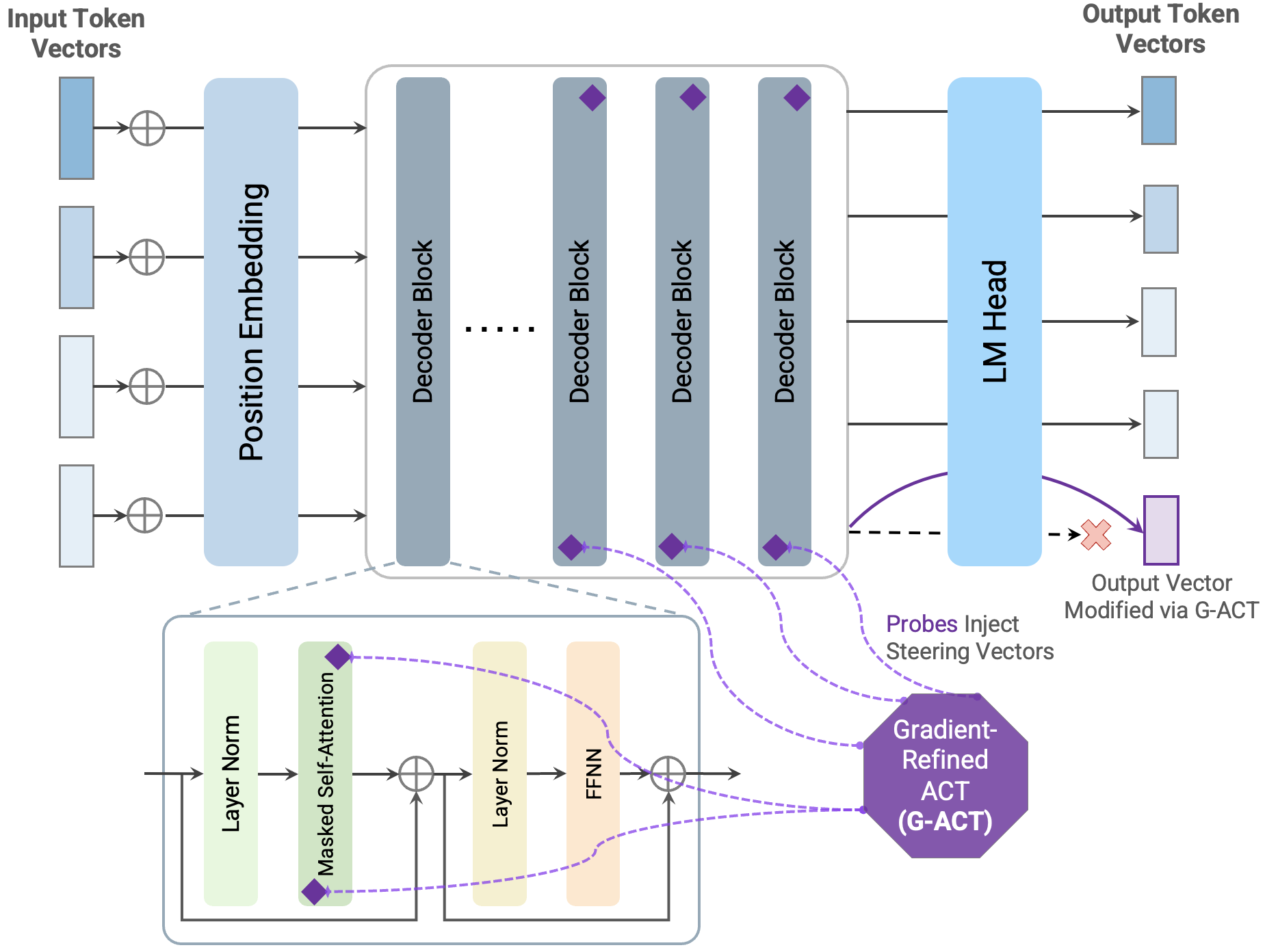}
  \caption{Schematic of Gradient-Refined ACT (G-ACT). Small per-layer probes tap each decoder block’s residual (post-attention or post-FFN) to predict a steering-vector cluster, which is then injected back into that same residual stream—biasing the model toward the target subject (CPP code) before the final LM head produces token logits.}
  \label{fig:G-ACT}
\end{figure}

The algorithm (\ref{alg:act-refine}) extends the original three‐step ACT framework \cite{act} by introducing a lightweight probe‐refinement stage. First, (1) per‑prompt style–difference vectors are extracted, (2) cluster these into a compact set of steering centroids, and (3) train a separate probe at each layer to assign new activations to these clusters at inference. In the second phase, probes are iteratively refined via gradient descent: autoregressive inference is performed under gradient tracking, and at each layer (a) the selected centroid as a residual adjustment is applied and (b) accumulate a cross‑entropy loss comparing the probe’s prediction to the known cluster label. Finally, the loss is propagated back exclusively through the probe parameters, keeping the weights of the base model fixed, as shown in Fig. \ref{fig:G-ACT}. This targeted refinement produces more accurate layer-specific classifiers and yields stronger and more consistent steering, while preserving the interpretability of the method and the minimal runtime cost.

\footnotesize{
\renewcommand{\thealgorithm}{}
\begin{algorithm}[hbt!] 
\caption{Gradient‑Refined Adaptive Activation Steering (G-ACT)}
\begin{algorithmic}[1]  \label{alg:act-refine}
\Require $\mathcal{M}$: pretrained language model  
\Require $\mathcal{D}=\{(Q_i,A_i^+,A_i^-)\}_{i=1}^N$: prompts with two reference answers  
\Require $C$: number of clusters  
\Require $E$: number of probe refinement epochs  
\Require $\alpha$: steering strength  
\Ensure Steered output $S$

\State Initialize empty list $V$  
\State Initialize empty list $P$  

\For{$i=1$ \textbf{to} $N$}
  \State Run $\mathcal{M}$ on $(Q_i,A_i^+)$, let $\mu_i^+=\mathrm{MeanAct}(\mathcal{M})$  
  \State Run $\mathcal{M}$ on $(Q_i,A_i^-)$, let $\mu_i^-=\mathrm{MeanAct}(\mathcal{M})$  
  \State Compute $v_i \gets \mu_i^+ - \mu_i^-$  
  \State Append $v_i$ to $V$
\EndFor

\State Partition $V$ into clusters $D_1,\dots,D_C \gets \mathrm{KMeans}(V)$  

\For{$j=1$ \textbf{to} $C$}
  \State $\pi_j \gets$ \Call{TrainProbe}{$D_j$}  
  \State Append $\pi_j$ to $P$
\EndFor

\Statex
\State \textbf{--- Probe Refinement ---}
\For{$e=1$ \textbf{to} $E$}
  \State Zero gradients of all probes in $P$  
  \State Set total loss $\mathcal{L}\gets 0$  
  \For{each prompt $(Q_i,A_i^+,A_i^-)\in\mathcal{D}$}
    \State Run a forward pass of $\mathcal{M}$ on $Q_i$, collecting per‐layer activations $\{h_{i,\ell}\}$  
    \For{$\ell=1$ \textbf{to} $L$}
      \State Compute logits $z_{i,\ell} \gets \pi_\ell(h_{i,\ell})$  
      \State $\mathcal{L}\mathrel{+}= \mathrm{CE}\bigl(z_{i,\ell},\,j_i\bigr)$  
      \State Inject steering: $h_{i,\ell}\leftarrow h_{i,\ell} + \alpha\,c_{j_i,\ell}$
    \EndFor
  \EndFor
  \State Backpropagate $\mathcal{L}$ and update probes in $P$
\EndFor

\State $S \gets \mathcal{M}.\mathrm{generate}(Q_{\mathrm{test}})$  
\end{algorithmic}
\end{algorithm} 
}

\noindent\textbf{Notations: } 
Let $\mathcal{M}$ be the pretrained language model to be steered, and let 
\[
  \mathcal{D} = \{(Q_i, A_i^+, A_i^-)\}_{i=1}^N
\]
denote the set of $N$ prompts $Q_i$, each paired with a target‐style answer $A_i^+$ and a baseline answer $A_i^-$.  We write $L$ for the number of transformer layers in $\mathcal{M}$, $C$ for the number of steering clusters discovered by K‑means, $E$ for the number of refinement epochs, and $\alpha$ for the scalar steering strength.  For prompt $i$ at layer $\ell$, let 
\[
  h_{i,\ell}^+,\;h_{i,\ell}^- \in \mathbb{R}^D
\]
be the activations under the two styles, and define the difference vector 
\[
  \Delta_{i,\ell} = h_{i,\ell}^+ - h_{i,\ell}^-.
\]
Cluster centroids are denoted 
\[
  \{c_{k}\}_{k=1}^C,\quad c_{k,\ell}\in\mathbb{R}^D,
\]
and each probe 
\[
  \pi_\ell:\mathbb{R}^D\to\mathbb{R}^C
\]
is a linear classifier that maps $h_{i,\ell}$ to logits 
\[
  z_{i,\ell} = \pi_\ell(h_{i,\ell}),
\]
used both to select the centroid $c_{k_i,\ell}$ and to compute the cross‐entropy loss $\mathrm{CE}(z_{i,\ell},k_i)$ against the true cluster label $k_i$.  Finally, $S$ denotes the steered output generated by $\mathcal{M}$ after all residual injections.

\subsection{Generating Dataset} \label{sec:datasetGen}
A dataset of coding challenges, inspired from the SciCode corpus \citep{tian2024scicoderesearchcodingbenchmark}, was assembled to isolate the natural language selection behavior of an LLM. Each example provides a problem statement (e.g., 'Implement the Gauss-Seidel method' or 'Generate the n$^\text{th}$ Fibonacci sequence') in addition to a problem description without mention of a programming language. The tasks span both general-purpose algorithms and domain-specific scientific routines, ensuring broad coverage as shown in the Snippet \ref{block:dataset}. Each problem instance in this dataset represents a distinct scientific or coding scenario; consequently, even after randomly partitioning into training and test subsets (e.g., an 80\%/20\% split), every held‑out example remains effectively out‑of‑distribution. In other words, there is no natural “in‑distribution” test set, since no two problems share identical underlying characteristics. Prompts are cast as direct questions to instruction‑tuned models, and qualitative comparisons of their outputs are used to assess shifts in chosen implementation language.

\begin{tcbraster}[raster valign=top, raster halign=center,raster columns=1, raster rows=1, colframe=white,colback=white,colbacktitle=teal!50!white,] \label{block:dataset}

\begin{tcolorbox}[title = Sample Task 1 - Problem Name: Gauss Seidel,colback=black!5!white,colframe=teal!75!black,fonttitle=\bfseries, width = 0.475\linewidt]
\footnotesize{Create a function to solve the matrix equation $Ax=b$ using the Gauss-Seidel iteration. The function takes a matrix $A$ and a vector $b$ as inputs. The method involves splitting the matrix $A$ into the difference of two matrices, $A=M-N$. For Gauss-Seidel, $M=D-L$, where $D$ is the diagonal component of $A$ and $L$ is the lower triangular component of $A$. The function should implement the corresponding iterative solvers until the norm of the increment is less than the given tolerance, $||x_k - x_{k-1}||_{l_2}<\ \epsilon$. 
}
\end{tcolorbox}

\begin{tcolorbox}[title = Sample Task 2 - Problem Name: Non-Uniform Meshing ,colback=black!5!white,colframe=teal!75!black,fonttitle=\bfseries, width = 0.475\linewidth]
\footnotesize{Write a code that automatically generates a one-dimensional non-uniform mesh with $n$ cells (grid points) between locations $y_0$ and $y_n$. The cell size $\Delta y_j = y_j - y_{j-1}$ varies according to a geometric progression: $\Delta y_{j+1} = \alpha \cdot \Delta y_j$ for $j = 1$ to $n-1$. Implement the mesh generation with $n = 256$ using a stretching factor $\alpha = 1.08$, between $y_0 = 0$ and $y_n = 1$. }
\end{tcolorbox}
\end{tcbraster} 


\section{Results} \label{results}
In this study, five instruction-tuned LLMs were evaluated:
Llama‑3.2‑3B‑Instruct, a compact 3 billion‑parameter model;
Llama‑3.3‑70B‑Instruct, its 70 billion‑parameter counterpart \citep{grattafiori2024llama3herdmodels};
Qwen2.5‑Coder‑32B‑Instruct, a 32 billion‑parameter model specialized for coding;
Qwen2.5‑14B‑Instruct‑1M, a 14 billion‑parameter variant fine‑tuned with a context window of one million tokens; and QwQ‑32B, a general‑purpose 32 billion‑parameter model \citep{qwen2025qwen25technicalreport}. Models were evaluated on the 84 benchmark questions at a sampling temperature (T) of 1.0. To verify statistical stability, each model–prompt pair was sampled both 100 and 25 times; observed performance differences between these two sampling sizes were under 1\%. Consequently, 25 repetitions per prompt were deemed sufficient and used for all reported results. 

\subsection{Language Preferences}
\begin{figure}[!htb]
  \centering
  \includegraphics[width=1.0\linewidth]{ 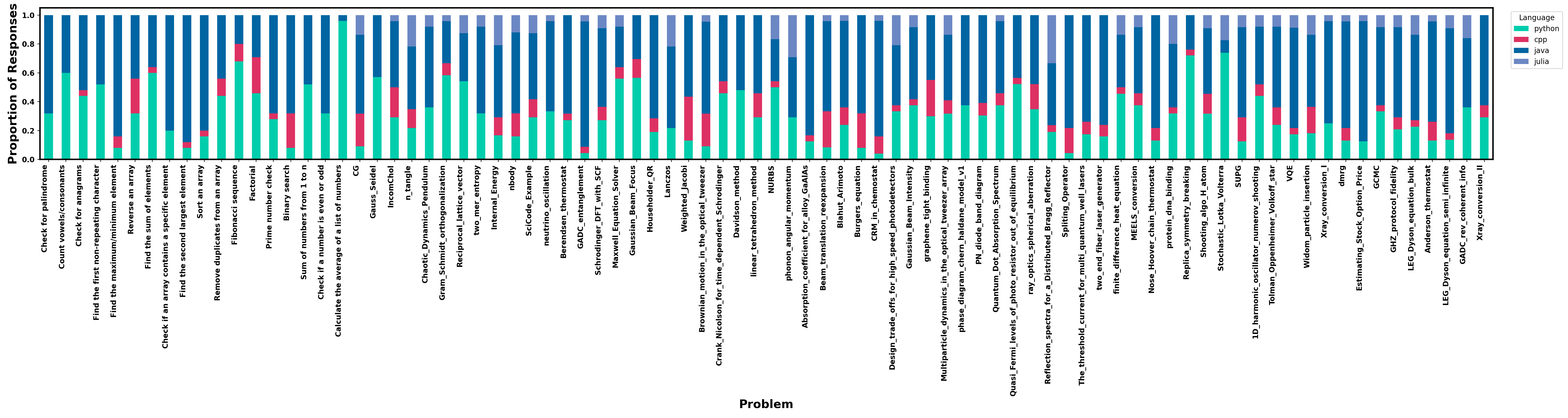}
  \caption{Programming‐language preferences of Llama‑3.2‑3B‑Instruct on a set of coding tasks (25 runs per task at T = 1.0).}
  \label{fig:ll3b}
\end{figure}

\begin{figure}[!htb]
  \centering
  \includegraphics[width=1.0\linewidth]{ 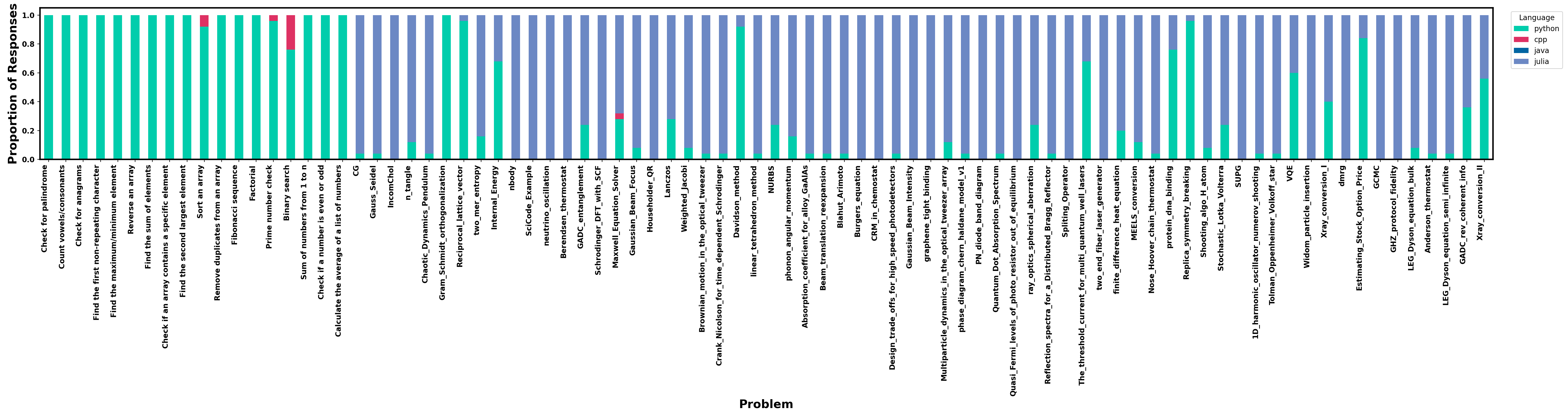}
  \caption{Programming‐language preferences of Qwen2.5-14B-Instruct-1M on a set of coding tasks (25 runs per task at T = 1.0).}
  \label{fig:Q14b}
\end{figure}
Figure \ref{fig:ll3b} illustrates the programming‑language distribution for Llama‑3.2‑3B‑Instruct across 84 benchmark problems. Python (green) predominates, accounting for roughly 70–80\% of outputs on nearly every task. Java (blue) appears most frequently on classical algorithmic challenges, such as sorting routines, string manipulations, and data structure implementations, peaking at about 30\%. CPP (red) is chosen only sporadically (<10\%), typically for numerically intensive or performance‑critical operations such as prime‑number checks and binary conversions. Julia (purple) applies exclusively to a small subset of domain-specific scientific problems (e.g. differential equation solvers, linear algebra benchmarks), also at low frequency (<10\%).
Figure \ref{fig:Q14b} presents the analogous results for Qwen2.5‑14B‑Instruct‑1M. In this case, Julia is the predominant output language ($\sim$ 67\%), with Python as the secondary preference ($\sim$ 33\%) and negligible use of Java or CPP. The results of the remaining models are presented in \ref{appendix:resutls_models}.

\begin{figure}[!htb]
  \centering
  \includegraphics[width=.75\linewidth]{ 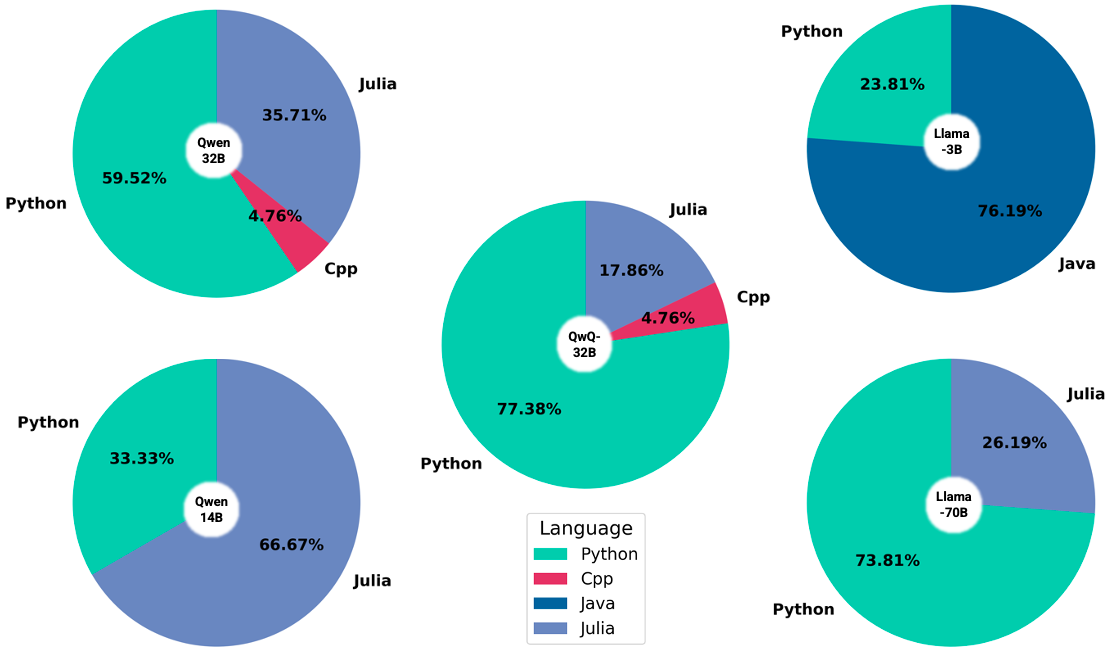}
  \caption{Aggregate programming‐language preferences of five LLMs on the coding benchmark (25 repetitions per task at sampling temperature 1.0). Each pie-chart shows the percentage of generated solutions in Python (green), CPP (red), Java (blue), and Julia (purple) for the corresponding model.}
  \label{fig:allRes}
\end{figure}

The summary of results for the five models is presented in Figure \ref{fig:allRes}. Llama‑3.2‑3B‑Instruct defaults strongly to Java (76.2\%) with the remainder in Python (23.8\%), whereas the larger Llama‑3.3‑70B‑Instruct model reverses that bias—Python accounts for 73.8\% of outputs and Julia for 26.2\%. Among the Qwen family, Qwen2.5‑Coder‑32B‑Instruct generates Python most often (59.5\%), followed by Julia (35.7\%) and a small 4.8\% share of CPP; Qwen2.5‑14B‑Instruct‑1M exhibits a pronounced Julia preference (66.7\%) over Python (33.3\%); and the general‑purpose QwQ‑32B again favors Python (77.4\%) with Julia at 17.9\% and CPP at 4.8\%. These findings illustrate that variations in model scale, architectural design, and fine‑tuning data collectively impart distinct, reproducible biases in each model’s code generation behavior. In particular, smaller, distilled models are often fine-tuned on corpora focused on web development and other routine applications. Based on these findings, to bias code generation towards the CPP language for scientific computing, there is a need for targeted alignment, through prompt engineering or domain-specific fine-tuning of the model.


\subsection{Static Method Analysis}
\begin{figure}[!htb]
  \centering
  \includegraphics[width=1.0\linewidth]{ 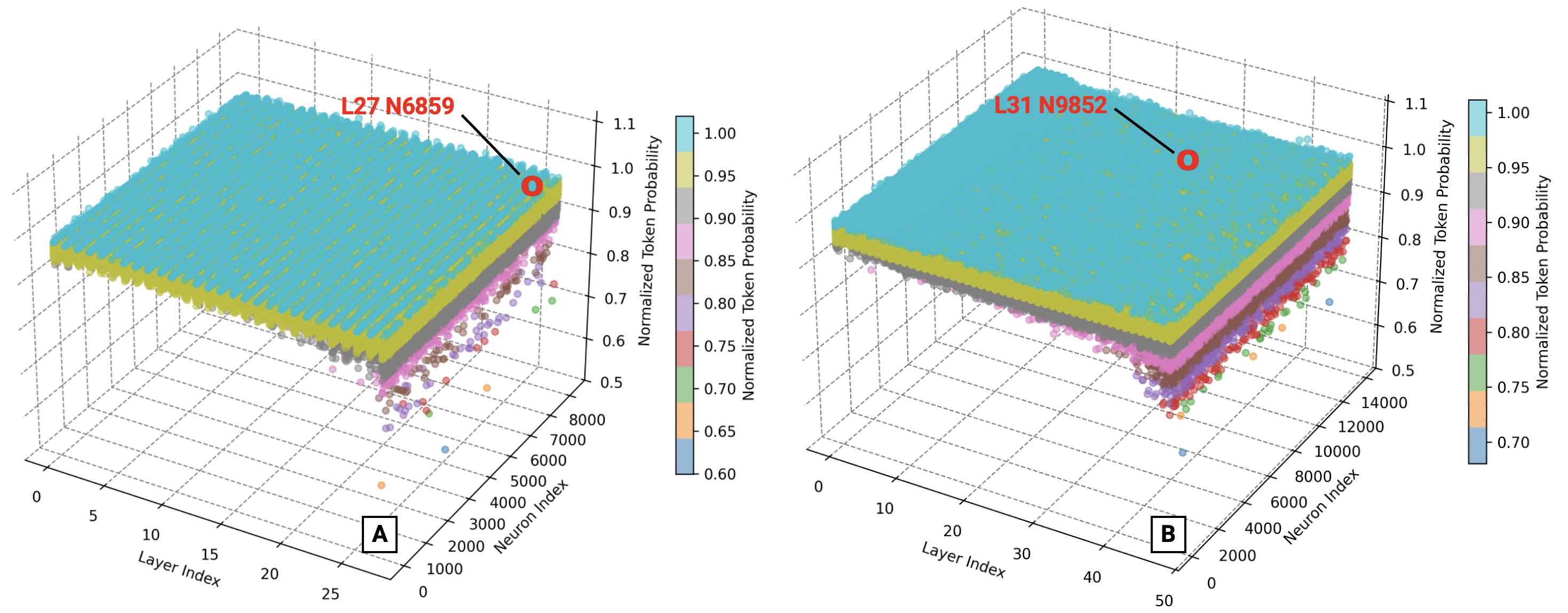}
  \caption{Activation maps for A. Llama‑3.2‑3B‑Instruct and B. Qwen2.5‑14B‑Instruct‑1M models respectively. Neuron with highest activation probability is marked with a red circle for CPP token.}
  \label{fig:3dmaps1}
\end{figure}

The procedure outlined in Sec. \ref{sec:mlp_static} was applied across all models. Figure \ref{fig:3dmaps1} shows the activation maps for Llama-3.2-3B-Instruct and Qwen2.5-14B-Instruct-1M, with the neuron exhibiting the highest probability for the “CPP” token located in layer 27 of the Llama model and layer 31 of the Qwen model. Activation maps for the remaining three models are shown in Figure \ref{fig:3dmaps2}.
\subsubsection{Inducing Language Preference Bias}
\begin{figure}[!htb]
  \centering
  \includegraphics[width=1.0\linewidth]{ 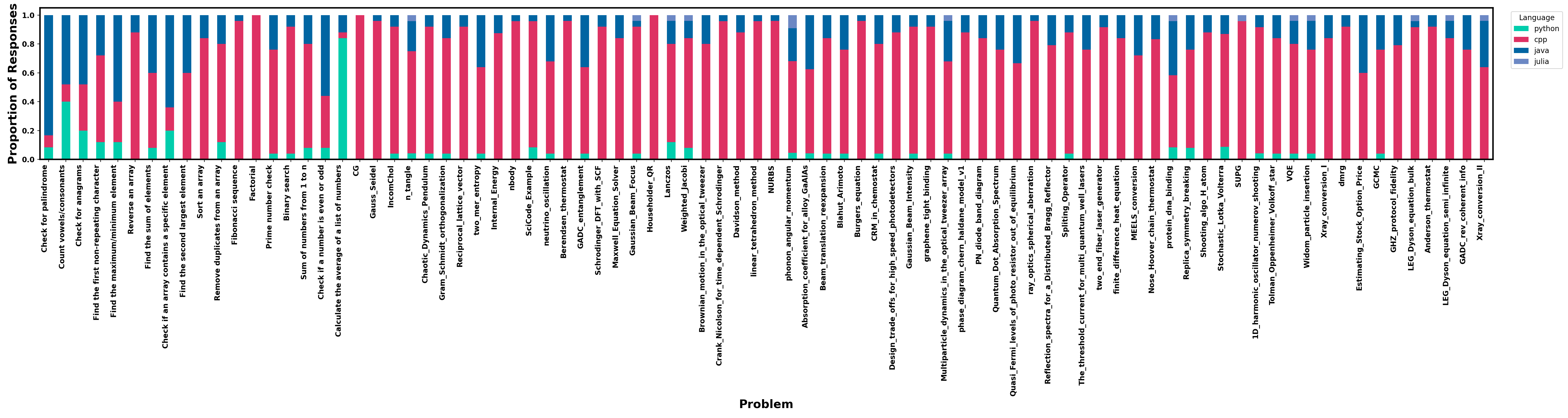}
  \caption{Programming‐language preferences of Llama‑3.2‑3B‑Instruct on a set of coding tasks (25 runs per task at T = 1.0) with activated neuron for CPP.}
  \label{fig:ll3b_cpp}
\end{figure}

To evaluate the causal role of the identified ``CPP'' neuron, Llama-3.2-3B-Instruct was rerun on the benchmark with artificially amplified activation of the same neuron (L27 N6859). The resulting shift in language selection frequencies, demonstrating an increase in CPP output, is shown in Fig. \ref{fig:ll3b_cpp}.

\begin{figure} [!htb]
  \centering
  \includegraphics[width=1.0\linewidth]{ 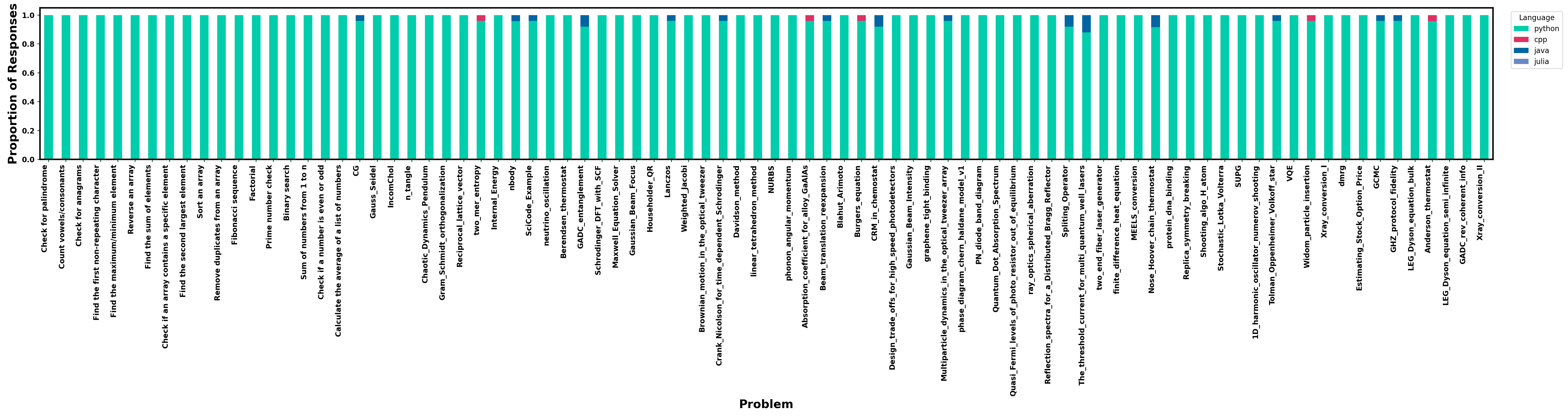}
  \caption{Programming‐language preferences of Llama‑3.2‑3B‑Instruct on a set of coding tasks (25 runs per task at T = 1.0) with activated neuron for Python.}
  \label{fig:ll3b_python}
\end{figure}

In the baseline configuration, Java was the dominant output language for Llama-3.2-3B-Instruct.
Compared to baseline (Fig. \ref{fig:ll3b}), where CPP never exceeded ~10 \%, the red bars now dominate almost all problems, often approaching or reaching the output of CPP 100 \%. Python (green) and Java (blue) virtually disappear, and Julia (purple) is completely eliminated. 
In a converse experiment, by applying the same algorithm to amplify the neuron associated with the ``Python'' token, the model was re-evaluated on the identical 84 task benchmark. As shown in Fig. \ref{fig:ll3b_python}, this single neuron drives the model to select Python for virtually every prompt, green bars reach nearly 100 \%, while CPP, Java and Julia outputs drop to negligible levels. These complementary tests confirm that selective activation of individual MLP neurons exerts a strong causal control over the model’s programming‐language choice, effectively overriding its native bias. The next section examines whether selective activation of the identified neuron can causally enforce a CPP-only code generation regime, irrespective of the original task context. 

\subsubsection{Code Generation Testing}
\begin{figure}[!htb]
  \centering
  \includegraphics[width=1.0\linewidth]{ 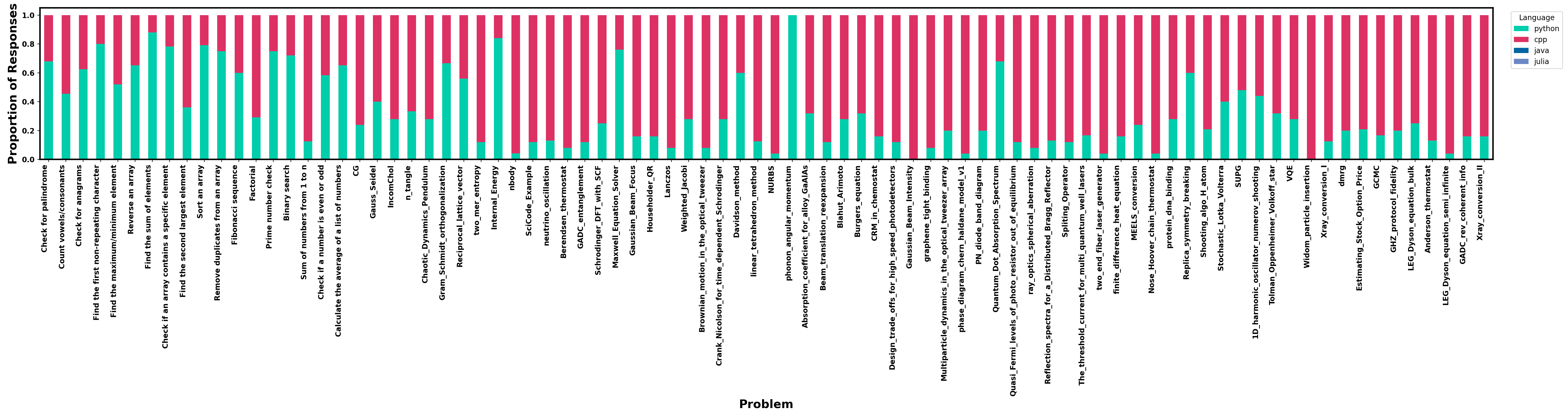}
  \caption{Code output of Llama‑3.2‑3B‑Instruct on a set of coding tasks (25 runs per task at T = 1.0) without activated neuron. }
  \label{fig:ll3b_Code}
\end{figure}

To establish a baseline, the Llama-3.2-3B-Instruct model was tasked (prompt in Sec.\ref{sec:prompts}) with generating code for each problem in the dataset by restricting its language choice exclusively to CPP and Python. The first two tokens of the generated output contain the markdown style format of the code: \text{```python} or \text{```cpp} that indicate the code preference. The overall results are shown in Fig. \ref{fig:ll3b_Code}, where two distinct regimes emerge. First, Python‐leaning tasks (e.g., high-level array manipulations, string processing) see Python outputs ranging from 40–80 \%. Second, CPP-dominant tasks (e.g., performance-critical routines, bitwise operations, numeric solvers) exhibit CPP output exceeding 60–90 \%.
Across all problems, the model defaults to CPP ($\sim$ 73 \%) more often than Python, but still resorts to Python (about 26 \%) for a substantial minority of prompts. This variability reflects the native bias of the model towards low-level implementations on computationally intensive tasks, while retaining flexibility to generate Python for simpler algorithmic challenges.

\begin{figure} [!htb]
  \centering
  \includegraphics[width=1.0\linewidth]{ 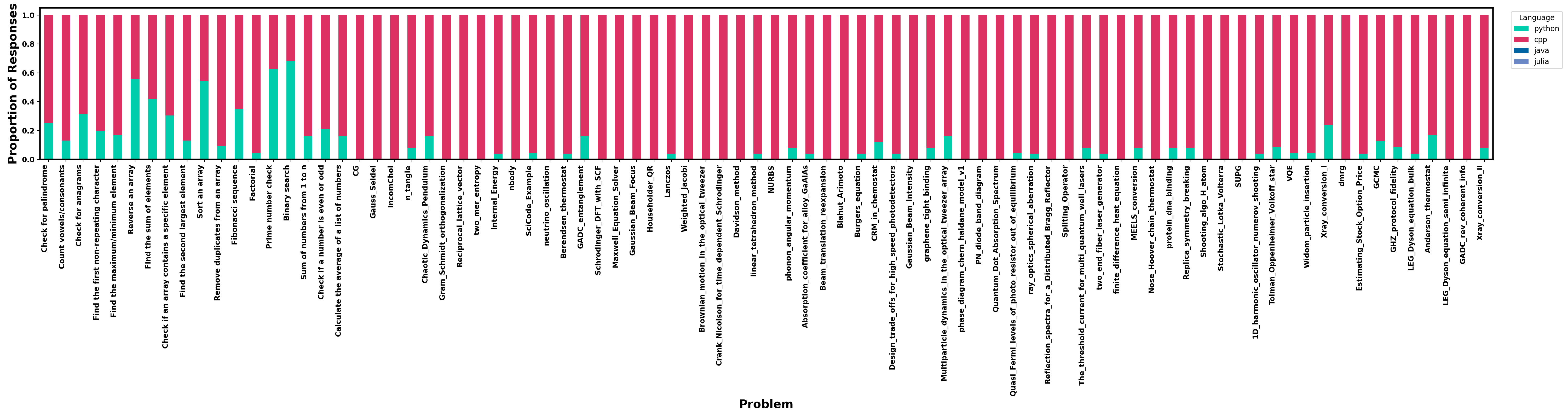}
  \caption{Code output of Llama‑3.2‑3B‑Instruct on a set of coding tasks (25 runs per task at T = 1.0) with activated neuron for CPP.}
  \label{fig:ll3b_codeCPP}
\end{figure}

\begin{tcbraster}[raster equal height,raster valign=top, raster halign=center,raster columns=2, raster rows=1,
colframe=white,colback=white,colbacktitle=teal!50!black,
]  \label{block:codeCPP}
\begin{tcolorbox}[title = Activated Neuron Llama‑3.2‑3B Model, colback=black!5!white,colframe=teal!75!black,fonttitle=\bfseries, height = 12.5cm]
\scriptsize{ 
\begin{verbatim}
#include <iostream>
#include <vector>
#include <cmath>
using namespace std;

void gauss_seidel(
const vector<vector<double>>& A, 
vector<double>& b, 
double tolerance, int max_iter) {
  int n = A.size();
  vector<double> x(n);
  for (int i = 0; i < n; i++) {
    for (int k = 0; k < n; k++) {
      double sum = 0.0;
      for (int j = 0; j < i; j++) {
        sum += A[i][j] * x[j];
      }
    }
    x[i] = (b[i] - sum) / A[i][i];
  }
  for (int iter = 0; 
  iter < max_iter; iter++) {
    vector<double> temp = x;
    for (int i = 0; i < n; i++) {
      double sum = 0.0;
      for (int k = 0; k < n; k++) {
        if (i!= k) {
          sum += A[i][k] * temp[k];
        }
      }
      x[i] = (b[i] - sum) / A[i][i];
    }
    double norm_x = 0.0;
    for (int i = 0; i < n; i++) {
      norm_x += pow(x[i]-temp[i], 2);
    }
    norm_x = sqrt(norm_x);
    if (norm_x < tolerance) break;
  }
}
\end{verbatim}
}
\end{tcolorbox}
\begin{tcolorbox}[boxsep=0pt,boxrule=0pt,colback=white,colframe=white,enhanced jigsaw,left=0mm,right=0mm,top=0mm,bottom=0mm]
\begin{tcolorbox}[title = Standard Llama‑3.2‑3B Model, colback=black!5!white,colframe=teal!75!black,fonttitle=\bfseries, height = 12.5cm]
\scriptsize{
\begin{verbatim}
def gauss_seidel(A, b, 
tolerance=1e-10, max_iter=1000):
n = len(b)
x = [0.0 for i in range(n)]
for i in range(n):
    x[i] = b[i]
    for j in range(i):
        x[i] -= A[i][j] * x[j]
    x[i] /= A[i][i]
for iter in range(max_iter):
    temp = [0.0 for i in range(n)]
    for i in range(n):
        temp[i] = x[i]
        for j in range(i):
            temp[i] -= A[i][j] * x[j]
        temp[i] /= A[i][i]
    norm_x = sum([pow(x_i - t_i, 2) 
    for x_i, t_i in zip(x, temp)])
    norm_x = pow(norm_x, 0.5)
    if norm_x < tolerance:
        break
    x = temp

return x

# Example usage
A = [[3, 2, 1], [2, 4, 1], [1, 1, 6]]
b = [8, 12, 20]
x = gauss_seidel(A, b)
print(x)
\end{verbatim}
}

\end{tcolorbox}
\end{tcolorbox}
\end{tcbraster}

For the identical prompt, activating the ``CPP'' neuron produces an overwhelmingly CPP‐centric output profile (red) as shown in Fig. \ref{fig:ll3b_codeCPP}. In almost every problem, CPP now constitutes about 80-100 \% of the 25 samples, while Python (green) retreats to at most 10–20 \% on a handful of tasks. This confirms that up‐weighting the identified unit exerts a causal, near‐complete override of the model’s native language preferences, driving it to generate CPP code almost exclusively, while a small residual Python ($\sim$ 5\%) tail on a few prompts suggests minimal contextual resistance. For brevity, the code generated by the model for the Gauss-Seidel problem is provided in Snippet \ref{block:codeCPP}. While the generated CPP implementation allocates the solution array, performs relaxation updates in nested loops, monitors convergence against the prescribed tolerance, and terminates once the residual falls below the threshold, a comprehensive validation of its correctness and performance has not been conducted and is beyond the scope of this study. Additionally, the model tends to generate Python code for the ``photon angular momentum'' problem in Fig. \ref{fig:ll3b_Code} that shifted to CPP after neuron activation. The generated code is shown in Snippet \ref{block:photonCode}. This shift from Python to CPP without modifying the prompt demonstrates that neuron-level activation can potentially override the model’s language choice while maintaining functional correctness.

\paragraph{What is the activation value for a single neuron?} All experiments so far apply a fixed perturbation to a single neuron to induce programming‐language bias. However, cross‐model evaluations reveal that both the optimal perturbation magnitude and the most effective neuron are strongly dependent on the model architecture and prompt phrasing. In addition to LM-head, activating neurons within MLP or attention‐head sublayers incurs minimal memory and computation overheads and is thus recommended \citep{kramár2024atpefficientscalablemethod}. We advise systematically tuning the perturbation amplitude and, where feasible, testing multiple candidate neurons, to identify an optimal configuration for any given prompt and model architecture.


\subsection{Adaptive Method Analysis}
To build the framework, the initial step involves splitting (70/30\%) the dataset of the coding problem prompts \ref{sec:datasetGen} into disjoint training and test sets. For each prompt $i$ in training set, two forward passes through the frozen model (here Llama‑3.2‑3B‑Instruct) are performed, one requesting a CPP solution and one requesting a Python solution, and the final token activations $h_{i,\ell}^+$ and $h_{i,\ell}^-$ at each layer $\ell$ are stored.  The per‐layer difference ($\Delta_{i,\ell}$) vectors were then used to characterize the "style shift" between CPP and Python, following the original ACT concept \citep{act}.

\begin{figure} [!htb]
  \centering
  \includegraphics[width=1.0\linewidth]{ 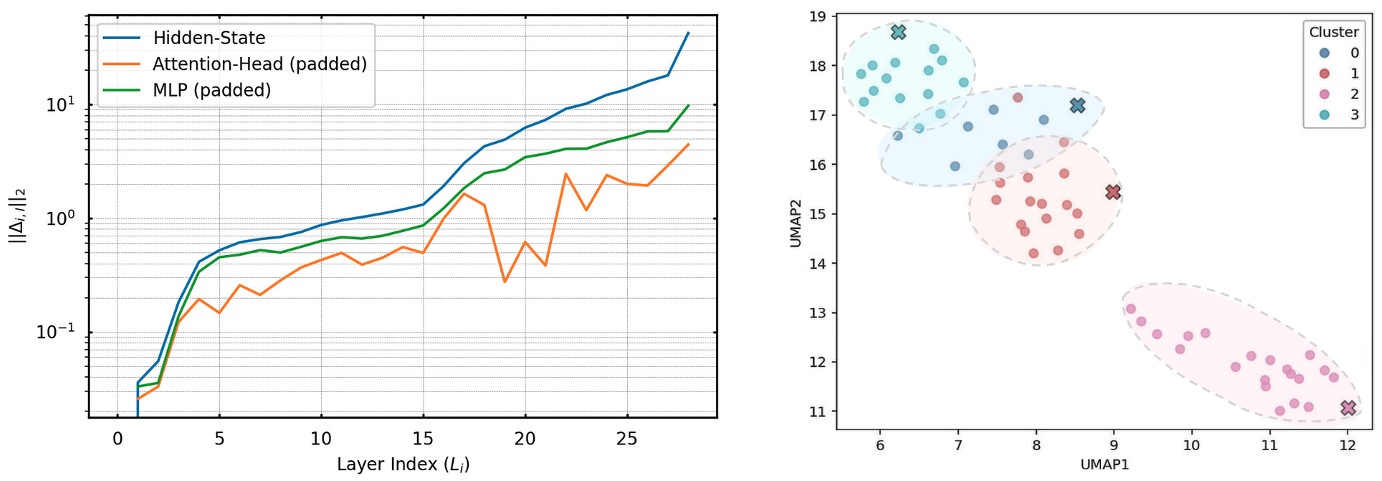}
  \caption{Results for Llama‑3.2‑3B‑Instruct model. Left: mean $\ell_2$‑norm of per‐layer CPP–Python difference vectors as a function of layer index. Right: UMAP projection of the flattened head‑difference vectors, colored by four K‑means clusters; crosses mark the cluster centroids.}
  \label{fig:diff-norm-clusters}
\end{figure}

In Figure~\ref{fig:diff-norm-clusters}, left plot shows the mean \(\ell_2\) norm of the vectors ($\|\Delta_{\ell}\|_2$) as a function of the layer index ($L_i$).  It is observed that the style signal is negligible in the earliest layers but grows steadily, reaching its maximum in the final blocks, which guided the decision to focus the steering efforts on the later layers.  While the full hidden‐state differences (blue) carry the largest absolute style signal in late layers, the attention‐head differences (orange) already exhibit a clear and much lower dimensional rise beginning around layer 5. Therefore, head outputs were selected to probe because they strike a practical balance between signal strength, interpretability, and computational cost: probing the entire hidden state or MLP output would require fitting and storing far larger clusters (and risks overfitting on spurious components of the style vector).

The right plot in figure~\ref{fig:diff-norm-clusters} visualizes the head‐difference vectors, in particular their four distinct modes when projected via UMAP \citep{mcinnes2018umap} and clustered by K‑means \citep{kmeans}. Each cluster centroid (cross) represents a family of prompts that induce very similar shifts (e.g., algorithmic or I/O intense problems) in the attention‑head activations when input switches between CPP and Python. Few outliers form their own small clusters because their style change is qualitatively different from the bulk of the tasks. 
Cluster 2 comprises about 20 straightforward coding tasks such as binary search, removing duplicates, array sorting, etc. which sit far from the other clusters. Cluster 3 (upper left) groups more complex algorithmic problems, while Cluster 1 contains prompts with descriptions of scientific variables (values, units) rather than pure code tasks. Although users may choose to cluster hidden‐state or MLP $\Delta_{i,\ell}$ vectors instead, the attention-head clusters offer a compact set of steering directions for the inference step.

\subsubsection{Per-Layer Activation Probes}
Per‑layer probe results for biasing generation toward CPP style are presented in this section. The standard ACT and the gradient-based ACT algorithms train their respective probes to optimize different objectives under different regimes. ACT fits a separate logistic regression on the flattened diffs of each layer, minimizing CE directly on that static dataset until convergence. The gradient-refined ACT embeds the probes inside an autoregressive generation loop: Each probe is updated by summing CE losses across every token step, while also injecting its chosen centroid back into the model. Here, the gradient-refined probes are trained for 50 epochs. 

Both sets of probes are evaluated on the testing set of prompts, which due to their uniqueness can be considered out-of-distribution (see Sec.\ref{sec:datasetGen}). To mitigate sensitivity to prompt phrasing, each test problem is evaluated using ten distinct CPP templates (e.g., "Problem: $\ldots$", "Implement the following in CPP: $\ldots$", etc.) and ten analogous Python templates. This ensemble of paraphrases ensures that the performance metrics reflect true style shifts rather than idiosyncrasies of any single prompt formulation.  

\paragraph{Evaluation Metrics} For each prompt and each transformer layer, the $\Delta_{i,\ell}$ vector is computed for every template and averaged, thus smoothing out surface-level variability, and then the probe predicts its nearest centroid. The probe classification accuracy and macro‑F1 scores serve as the comparison metric. This comparison method provides a clear, layer‐wise view of which method more reliably recognizes the correct steering mode by isolating probe quality from downstream generation effects. 

\textbf{Accuracy:} Fraction of test prompts for which the per-layer probe correctly predicts the cluster ID (assigned during training). Given $N$ test examples with true labels $y_i$ and predicted labels $\hat y_i$, accuracy is $\mathrm{Acc} = \frac{1}{N}\sum_{i=1}^N \mathbf{1}\!\bigl(\hat y_i = y_i\bigr)$, where $\mathbf{1}(\cdot)$ is the indicator function.

\textbf{Macro‑F1:} Unweighted average of the F1 scores across all clusters, giving equal importance to small and large clusters. For a $C$-class problem, let $\mathrm{TP}_c,\ \mathrm{FP}_c,\ \mathrm{FN}_c$ be the true‐positive, false‐positive, and false‐negative counts for class $c$. Define per‐class precision, recall, and F1 as
\[
  \mathrm{Precision}_c = \frac{\mathrm{TP}_c}{\mathrm{TP}_c + \mathrm{FP}_c}, 
  \quad
  \mathrm{Recall}_c    = \frac{\mathrm{TP}_c}{\mathrm{TP}_c + \mathrm{FN}_c},
  \quad
  \mathrm{F1}_c        = \frac{2\,\mathrm{Precision}_c\,\mathrm{Recall}_c}{\mathrm{Precision}_c + \mathrm{Recall}_c}.
\]
As mentioned earlier, Macro‑F1 is then the unweighted mean of the per‐class F1 scores: $\mathrm{Macro\text{-}F1} =  \frac{1}{C}\sum_{c=1}^{C} \mathrm{F1}_c$.

\begin{figure} [!htb]
  \centering
  \includegraphics[width=1.0\linewidth]{ 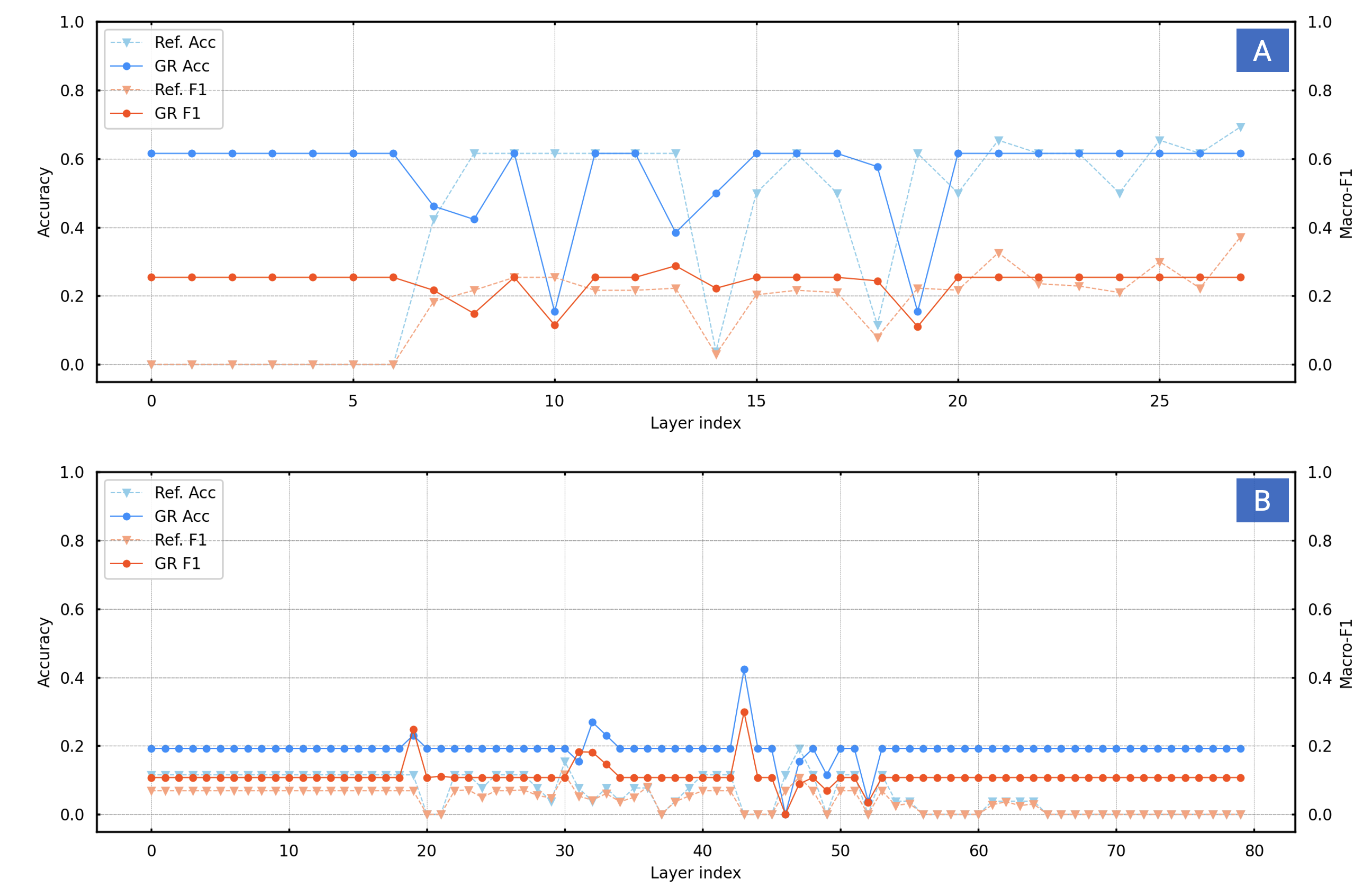}
  \caption{Results for A. Llama‑3.2‑3B‑Instruct model and B. Llama‑3.3‑70B‑Instruct model. $Acc_{\mathrm{Ref}}$ and $Acc_{\mathrm{GR}}$ denote the proportion of prompts correctly assigned to their nearest centroid by the standard and refined probes, respectively. $F1_{\mathrm{Ref}}$ and $F1_{\mathrm{GR}}$ are the corresponding macro‐averaged F1 scores.}
  \label{fig:ood-llamaModels}
\end{figure}

In Fig.\ref{fig:ood-llamaModels}, the gradient-refined ACT probes deliver strong improvements on Llama 3B but collapse when applied to Llama 70B. For the 3B model, gradient-based refinement increases the accuracy of the early layers 0–6 from 0\% to 61. 5\% (and 0 to 0.254 macroF1), improving what were previously uninformative probes. Midlayers (7, 11-12, 15-17, 20, 22-27) see modest to significant gains, while a few layers (8, 10, 13, 19) tend to under-predict with refinement, suggesting overfitting or class collapse in those specific blocks. Layers that were already strong in standard ACT (9, 16, 21, 23, 25, 27) remain unchanged at peak performance. Averaged across all 28 layers, the accuracy increases from 0.405 to 0.556 (+15\%) and the macro‑F1 from 0.165 to 0.238 (+7.3\%), demonstrating that the gradient-refined ACT probes more reliably recognize the correct steering mode, especially in layers that originally carried a smaller or noisy style signal. In contrast, on the much larger 70B model, where only hidden-state differences were used and clusters reduced to three (see Fig. \ref{fig:diff-norm-clusters_70B}), both standard ACT and gradient ACT probes underperform: the mean accuracy is only 6.3\% for vanilla and 19.1\% for refined; mean macro-F1 is 0.041 vs. 0.111. In other words, the head-state activations that carry a clear style signal in 3B become too diffuse or noisy in 70B model, so clustering and probes trained on those attention-head features lose their discriminative power; instead hidden-state features (see Fig. \ref{fig:diff-norm-clusters}) can be used in such scenarios. 

\begin{table}[!htb]
  \centering
  \begin{tabular}{lcc}
    \toprule
    & \multicolumn{2}{c}{\textbf{Model}} \\
    \cmidrule(lr){2-3}
    \textbf{Method}           & \textbf{Llama‑3.2‑3B‑Instruct} & \textbf{Llama‑3.3‑70B‑Instruct} \\
    \midrule
    Vanilla Model          & 3.2502 $\pm$ 0.0033 s   & 27.7948 $\pm$ 0.0025 s   \\
    Standard ACT           & 6.7840 $\pm$ 0.0049 s   & 38.7461 $\pm$ 0.0177 s   \\
    Gradient-Refined ACT   & 4.5526 $\pm$ 0.0367 s   & 32.6276 $\pm$ 0.0965 s  \\
    \bottomrule
  \end{tabular}
  \caption{Inference time performance of per-layer probes under standard and gradient-based ACT methods compared to no-probe setting for different Llama family models. Average timings are reported across 25 inference runs, after discarding the first 5 warm‐up iterations.}
  \label{tab:llama-act-comparison}
\end{table}

Furthermore, the probe logic does come with a non-trivial runtime cost. As shown in Table \ref{tab:llama-act-comparison}, on Llama-3B a single ``no-probe" forward pass takes approx. 3.25 s, but inserting standard ACT injections inflates that to about 6.78 s (nearly double). Gradient-refined ACT falls in between ($\sim$ 4.55 s) as it must evaluate the lightweight classifier of each layer during generation but does not rerun the clustering offline. The 70B model shows the same pattern at scale: base generation costs about 27.8 s, standard ACT around 38.7 s, and gradient-refined ACT takes 32.6 s. In practice, each per-layer probe entails a small extra linear pass (and, when using gradients, a backward step), so total latency increases compared to unmodified inference.  Gradient-ACT skips the expensive nearest-centroid search at every layer. In standard ACT, $\Delta_{i,\ell}$ of each prompt token must be compared ($L2$ distance) with all K centroids to select which steering vector to add, resulting in an $\mathcal{O}(KD)$ operation per layer per token. In contrast, gradient-ACT uses a smaller linear probe (single matrix multiplication) to predict the cluster, so it avoids repeatedly computing distances. As a result, even though gradient-ACT still evaluates one extra linear layer per token, it ends up faster than a full K-means lookup at inference time, and hence the intermediate runtime.

\section{Conclusion}
Five causal language models were evaluated on a suite of scientific coding prompts to quantify coding language bias. While smaller models preferred Java or Julia inconsistently, both variants of the LLaMA models exhibited clear separability in their internal activations when prompted to select one of four language options.

A static neuron‐attribution approach, identifying individual LM-head neurons correlated with the CPP style and manually perturbing them, is shown to produce CPP code in limited cases. However, this method proved fragile, with performance highly sensitive to choice of neuron, prompt formulation, and model scale. To overcome these limitations, a gradient‐refined version of ACT is introduced. Per-prompt ``difference" vectors (CPP minus Python activations) are clustered into a small set of representative steering directions, and lightweight per-layer probes are trained, then further refined online via cross-entropy loss, during generation. In LLaMA-3.2 3B, the average probe classification accuracy of the probe increased, with the layers (0–6) improving from 0 \% to 61.5 \% accuracy. Even on the larger LLaMA-3.3 70B model, where head‐state signals become more diffuse, G-ACT still increases accuracy, demonstrating that targeted injections at key layers can reliably bias generation toward CPP despite overall weak activations.

Although adding per-layer probes incurs a modest runtime overhead (approximately 1.3–1.4× slower than base generation), practical deployments can accommodate this cost by steering only a subset of layers, caching subtokens, or accepting slightly longer latencies. Even imperfect probes suffice to steer the model output toward the desired style, yielding substantial qualitative gains in CPP code generation, and can be extended to other concepts of interest.  
While static neuron perturbation may only be viable in narrow scenarios, G-ACT provides a scalable, interpretable, and efficient mechanism for steering LLMs toward CPP (or any target subject), with acceptable inference costs. Beyond steering concepts, this approach embeds persistent transformation matrices that guarantee identical model behavior across different users, fostering a new paradigm of output reproducibility.

\begin{ack}
The authors acknowledge support from Los Alamos National Laboratory under the project ``Algorithm/Software/Hardware Co-design for High Energy Density Applications" at the University of Michigan. 
\end{ack}

\newpage
\bibliographystyle{plainnat}
\bibliography{citations}

\begin{thebibliography}{45}
\providecommand{\natexlab}[1]{#1}
\providecommand{\url}[1]{\texttt{#1}}
\expandafter\ifx\csname urlstyle\endcsname\relax
  \providecommand{\doi}[1]{doi: #1}\else
  \providecommand{\doi}{doi: \begingroup \urlstyle{rm}\Url}\fi

\bibitem[Arthur and Vassilvitskii(2007)]{kmeans}
David Arthur and Sergei Vassilvitskii.
\newblock k-means++: the advantages of careful seeding.
\newblock In \emph{Proceedings of the Eighteenth Annual ACM-SIAM Symposium on Discrete Algorithms}, SODA '07, page 1027–1035, USA, 2007. Society for Industrial and Applied Mathematics.
\newblock ISBN 9780898716245.

\bibitem[Brown et~al.(2023)Brown, Godfrey, Nizinski, Tu, and Kvinge]{brown2023robustness}
Davis Brown, Charles Godfrey, Cody Nizinski, Jonathan Tu, and Henry Kvinge.
\newblock Robustness of edited neural networks.
\newblock In \emph{ICLR 2023 Workshop on Mathematical and Empirical Understanding of Foundation Models}, 2023.

\bibitem[Chen et~al.(2024)Chen, Huang, Zheng, Du, and Zhang]{chen2024robust}
Jinyin Chen, Guohan Huang, Haibin Zheng, Hang Du, and Jian Zhang.
\newblock Robust explanations for graph neural network with neuron explanation component.
\newblock \emph{Information Sciences}, 654:\penalty0 119785, 2024.

\bibitem[Condori and Bruno(2021)]{condori2021analysis}
Rayner~HM Condori and Odemir~M Bruno.
\newblock Analysis of activation maps through global pooling measurements for texture classification.
\newblock \emph{Information Sciences}, 555:\penalty0 260--279, 2021.

\bibitem[Dathathri et~al.(2019)Dathathri, Madotto, Lan, Hung, Frank, Molino, Yosinski, and Liu]{dathathri2019plug}
Sumanth Dathathri, Andrea Madotto, Janice Lan, Jane Hung, Eric Frank, Piero Molino, Jason Yosinski, and Rosanne Liu.
\newblock Plug and play language models: A simple approach to controlled text generation.
\newblock \emph{arXiv preprint arXiv:1912.02164}, 2019.

\bibitem[Davies(2025)]{davies2025flashinterpretability}
Harry~J Davies.
\newblock Flash interpretability: Decoding specialised feature neurons in large language models with the lm-head, 2025.
\newblock URL \url{https://arxiv.org/abs/2501.02688}.

\bibitem[Elhage et~al.(2022)]{anthropicBasisToy}
Nelson Elhage et~al.
\newblock {Toy Models of Superposition}.
\newblock \emph{Transformer Circuits Thread}, 2022.
\newblock URL \url{https://transformer-circuits.pub/2022/toy_model/index.html#motivation-privileged}.

\bibitem[Gao et~al.(2024)Gao, Fang, Huang, Giunchiglia, Noori, Schwarz, Ektefaie, Kondic, and Zitnik]{gao2024empowering}
Shanghua Gao, Ada Fang, Yepeng Huang, Valentina Giunchiglia, Ayush Noori, Jonathan~Richard Schwarz, Yasha Ektefaie, Jovana Kondic, and Marinka Zitnik.
\newblock Empowering biomedical discovery with ai agents.
\newblock \emph{Cell}, 187\penalty0 (22):\penalty0 6125--6151, 2024.

\bibitem[Geiger et~al.(2022)Geiger, Wu, Lu, Rozner, Kreiss, Icard, Goodman, and Potts]{geiger2022inducing}
Atticus Geiger, Zhengxuan Wu, Hanson Lu, Josh Rozner, Elisa Kreiss, Thomas Icard, Noah Goodman, and Christopher Potts.
\newblock Inducing causal structure for interpretable neural networks.
\newblock In \emph{International Conference on Machine Learning}, pages 7324--7338. PMLR, 2022.

\bibitem[Grattafiori et~al.(2024)Grattafiori, Dubey, Jauhri, Pandey, Kadian, Al-Dahle, Letman, Mathur, Schelten, Vaughan, Yang, Fan, Goyal, Hartshorn, Yang, Mitra, Sravankumar, Korenev, Hinsvark, Rao, Zhang, Rodriguez, Gregerson, Spataru, Roziere, Biron, Tang, Chern, Caucheteux, Nayak, Bi, Marra, McConnell, Keller, Touret, Wu, Wong, Ferrer, Nikolaidis, Allonsius, Song, Pintz, Livshits, Wyatt, Esiobu, Choudhary, Mahajan, Garcia-Olano, Perino, Hupkes, Lakomkin, AlBadawy, Lobanova, Dinan, Smith, Radenovic, Guzmán, Zhang, Synnaeve, Lee, Anderson, Thattai, Nail, Mialon, Pang, Cucurell, Nguyen, Korevaar, Xu, Touvron, Zarov, Ibarra, Kloumann, Misra, Evtimov, Zhang, Copet, Lee, Geffert, Vranes, Park, Mahadeokar, Shah, van~der Linde, Billock, Hong, Lee, Fu, Chi, Huang, Liu, Wang, Yu, Bitton, Spisak, Park, Rocca, Johnstun, Saxe, Jia, Alwala, Prasad, Upasani, Plawiak, Li, Heafield, Stone, El-Arini, Iyer, Malik, Chiu, Bhalla, Lakhotia, Rantala-Yeary, van~der Maaten, Chen, Tan, Jenkins, Martin, Madaan, Malo, Blecher,
  Landzaat, de~Oliveira, Muzzi, Pasupuleti, Singh, Paluri, Kardas, Tsimpoukelli, Oldham, Rita, Pavlova, Kambadur, Lewis, Si, Singh, Hassan, Goyal, Torabi, Bashlykov, Bogoychev, Chatterji, Zhang, Duchenne, Çelebi, Alrassy, Zhang, Li, Vasic, Weng, Bhargava, Dubal, Krishnan, Koura, Xu, He, Dong, Srinivasan, Ganapathy, Calderer, Cabral, Stojnic, Raileanu, Maheswari, Girdhar, Patel, Sauvestre, Polidoro, Sumbaly, Taylor, Silva, Hou, Wang, Hosseini, Chennabasappa, Singh, Bell, Kim, Edunov, Nie, Narang, Raparthy, Shen, Wan, Bhosale, Zhang, Vandenhende, Batra, Whitman, Sootla, Collot, Gururangan, Borodinsky, Herman, Fowler, Sheasha, Georgiou, Scialom, Speckbacher, Mihaylov, Xiao, Karn, Goswami, Gupta, Ramanathan, Kerkez, Gonguet, Do, Vogeti, Albiero, Petrovic, Chu, Xiong, Fu, Meers, Martinet, Wang, Wang, Tan, Xia, Xie, Jia, Wang, Goldschlag, Gaur, Babaei, Wen, Song, Zhang, Li, Mao, Coudert, Yan, Chen, Papakipos, Singh, Srivastava, Jain, Kelsey, Shajnfeld, Gangidi, Victoria, Goldstand, Menon, Sharma, Boesenberg,
  Baevski, Feinstein, Kallet, Sangani, Teo, Yunus, Lupu, Alvarado, Caples, Gu, Ho, Poulton, Ryan, Ramchandani, Dong, Franco, Goyal, Saraf, Chowdhury, Gabriel, Bharambe, Eisenman, Yazdan, James, Maurer, Leonhardi, Huang, Loyd, Paola, Paranjape, Liu, Wu, Ni, Hancock, Wasti, Spence, Stojkovic, Gamido, Montalvo, Parker, Burton, Mejia, Liu, Wang, Kim, Zhou, Hu, Chu, Cai, Tindal, Feichtenhofer, Gao, Civin, Beaty, Kreymer, Li, Adkins, Xu, Testuggine, David, Parikh, Liskovich, Foss, Wang, Le, Holland, Dowling, Jamil, Montgomery, Presani, Hahn, Wood, Le, Brinkman, Arcaute, Dunbar, Smothers, Sun, Kreuk, Tian, Kokkinos, Ozgenel, Caggioni, Kanayet, Seide, Florez, Schwarz, Badeer, Swee, Halpern, Herman, Sizov, Guangyi, Zhang, Lakshminarayanan, Inan, Shojanazeri, Zou, Wang, Zha, Habeeb, Rudolph, Suk, Aspegren, Goldman, Zhan, Damlaj, Molybog, Tufanov, Leontiadis, Veliche, Gat, Weissman, Geboski, Kohli, Lam, Asher, Gaya, Marcus, Tang, Chan, Zhen, Reizenstein, Teboul, Zhong, Jin, Yang, Cummings, Carvill, Shepard, McPhie,
  Torres, Ginsburg, Wang, Wu, U, Saxena, Khandelwal, Zand, Matosich, Veeraraghavan, Michelena, Li, Jagadeesh, Huang, Chawla, Huang, Chen, Garg, A, Silva, Bell, Zhang, Guo, Yu, Moshkovich, Wehrstedt, Khabsa, Avalani, Bhatt, Mankus, Hasson, Lennie, Reso, Groshev, Naumov, Lathi, Keneally, Liu, Seltzer, Valko, Restrepo, Patel, Vyatskov, Samvelyan, Clark, Macey, Wang, Hermoso, Metanat, Rastegari, Bansal, Santhanam, Parks, White, Bawa, Singhal, Egebo, Usunier, Mehta, Laptev, Dong, Cheng, Chernoguz, Hart, Salpekar, Kalinli, Kent, Parekh, Saab, Balaji, Rittner, Bontrager, Roux, Dollar, Zvyagina, Ratanchandani, Yuvraj, Liang, Alao, Rodriguez, Ayub, Murthy, Nayani, Mitra, Parthasarathy, Li, Hogan, Battey, Wang, Howes, Rinott, Mehta, Siby, Bondu, Datta, Chugh, Hunt, Dhillon, Sidorov, Pan, Mahajan, Verma, Yamamoto, Ramaswamy, Lindsay, Lindsay, Feng, Lin, Zha, Patil, Shankar, Zhang, Zhang, Wang, Agarwal, Sajuyigbe, Chintala, Max, Chen, Kehoe, Satterfield, Govindaprasad, Gupta, Deng, Cho, Virk, Subramanian, Choudhury,
  Goldman, Remez, Glaser, Best, Koehler, Robinson, Li, Zhang, Matthews, Chou, Shaked, Vontimitta, Ajayi, Montanez, Mohan, Kumar, Mangla, Ionescu, Poenaru, Mihailescu, Ivanov, Li, Wang, Jiang, Bouaziz, Constable, Tang, Wu, Wang, Wu, Gao, Kleinman, Chen, Hu, Jia, Qi, Li, Zhang, Zhang, Adi, Nam, Yu, Wang, Zhao, Hao, Qian, Li, He, Rait, DeVito, Rosnbrick, Wen, Yang, Zhao, and Ma]{grattafiori2024llama3herdmodels}
Aaron Grattafiori, Abhimanyu Dubey, Abhinav Jauhri, Abhinav Pandey, Abhishek Kadian, Ahmad Al-Dahle, Aiesha Letman, Akhil Mathur, Alan Schelten, Alex Vaughan, Amy Yang, Angela Fan, Anirudh Goyal, Anthony Hartshorn, Aobo Yang, Archi Mitra, Archie Sravankumar, Artem Korenev, Arthur Hinsvark, Arun Rao, Aston Zhang, Aurelien Rodriguez, Austen Gregerson, Ava Spataru, Baptiste Roziere, Bethany Biron, Binh Tang, Bobbie Chern, Charlotte Caucheteux, Chaya Nayak, Chloe Bi, Chris Marra, Chris McConnell, Christian Keller, Christophe Touret, Chunyang Wu, Corinne Wong, Cristian~Canton Ferrer, Cyrus Nikolaidis, Damien Allonsius, Daniel Song, Danielle Pintz, Danny Livshits, Danny Wyatt, David Esiobu, Dhruv Choudhary, Dhruv Mahajan, Diego Garcia-Olano, Diego Perino, Dieuwke Hupkes, Egor Lakomkin, Ehab AlBadawy, Elina Lobanova, Emily Dinan, Eric~Michael Smith, Filip Radenovic, Francisco Guzmán, Frank Zhang, Gabriel Synnaeve, Gabrielle Lee, Georgia~Lewis Anderson, Govind Thattai, Graeme Nail, Gregoire Mialon, Guan Pang,
  Guillem Cucurell, Hailey Nguyen, Hannah Korevaar, Hu~Xu, Hugo Touvron, Iliyan Zarov, Imanol~Arrieta Ibarra, Isabel Kloumann, Ishan Misra, Ivan Evtimov, Jack Zhang, Jade Copet, Jaewon Lee, Jan Geffert, Jana Vranes, Jason Park, Jay Mahadeokar, Jeet Shah, Jelmer van~der Linde, Jennifer Billock, Jenny Hong, Jenya Lee, Jeremy Fu, Jianfeng Chi, Jianyu Huang, Jiawen Liu, Jie Wang, Jiecao Yu, Joanna Bitton, Joe Spisak, Jongsoo Park, Joseph Rocca, Joshua Johnstun, Joshua Saxe, Junteng Jia, Kalyan~Vasuden Alwala, Karthik Prasad, Kartikeya Upasani, Kate Plawiak, Ke~Li, Kenneth Heafield, Kevin Stone, Khalid El-Arini, Krithika Iyer, Kshitiz Malik, Kuenley Chiu, Kunal Bhalla, Kushal Lakhotia, Lauren Rantala-Yeary, Laurens van~der Maaten, Lawrence Chen, Liang Tan, Liz Jenkins, Louis Martin, Lovish Madaan, Lubo Malo, Lukas Blecher, Lukas Landzaat, Luke de~Oliveira, Madeline Muzzi, Mahesh Pasupuleti, Mannat Singh, Manohar Paluri, Marcin Kardas, Maria Tsimpoukelli, Mathew Oldham, Mathieu Rita, Maya Pavlova, Melanie Kambadur,
  Mike Lewis, Min Si, Mitesh~Kumar Singh, Mona Hassan, Naman Goyal, Narjes Torabi, Nikolay Bashlykov, Nikolay Bogoychev, Niladri Chatterji, Ning Zhang, Olivier Duchenne, Onur Çelebi, Patrick Alrassy, Pengchuan Zhang, Pengwei Li, Petar Vasic, Peter Weng, Prajjwal Bhargava, Pratik Dubal, Praveen Krishnan, Punit~Singh Koura, Puxin Xu, Qing He, Qingxiao Dong, Ragavan Srinivasan, Raj Ganapathy, Ramon Calderer, Ricardo~Silveira Cabral, Robert Stojnic, Roberta Raileanu, Rohan Maheswari, Rohit Girdhar, Rohit Patel, Romain Sauvestre, Ronnie Polidoro, Roshan Sumbaly, Ross Taylor, Ruan Silva, Rui Hou, Rui Wang, Saghar Hosseini, Sahana Chennabasappa, Sanjay Singh, Sean Bell, Seohyun~Sonia Kim, Sergey Edunov, Shaoliang Nie, Sharan Narang, Sharath Raparthy, Sheng Shen, Shengye Wan, Shruti Bhosale, Shun Zhang, Simon Vandenhende, Soumya Batra, Spencer Whitman, Sten Sootla, Stephane Collot, Suchin Gururangan, Sydney Borodinsky, Tamar Herman, Tara Fowler, Tarek Sheasha, Thomas Georgiou, Thomas Scialom, Tobias Speckbacher,
  Todor Mihaylov, Tong Xiao, Ujjwal Karn, Vedanuj Goswami, Vibhor Gupta, Vignesh Ramanathan, Viktor Kerkez, Vincent Gonguet, Virginie Do, Vish Vogeti, Vítor Albiero, Vladan Petrovic, Weiwei Chu, Wenhan Xiong, Wenyin Fu, Whitney Meers, Xavier Martinet, Xiaodong Wang, Xiaofang Wang, Xiaoqing~Ellen Tan, Xide Xia, Xinfeng Xie, Xuchao Jia, Xuewei Wang, Yaelle Goldschlag, Yashesh Gaur, Yasmine Babaei, Yi~Wen, Yiwen Song, Yuchen Zhang, Yue Li, Yuning Mao, Zacharie~Delpierre Coudert, Zheng Yan, Zhengxing Chen, Zoe Papakipos, Aaditya Singh, Aayushi Srivastava, Abha Jain, Adam Kelsey, Adam Shajnfeld, Adithya Gangidi, Adolfo Victoria, Ahuva Goldstand, Ajay Menon, Ajay Sharma, Alex Boesenberg, Alexei Baevski, Allie Feinstein, Amanda Kallet, Amit Sangani, Amos Teo, Anam Yunus, Andrei Lupu, Andres Alvarado, Andrew Caples, Andrew Gu, Andrew Ho, Andrew Poulton, Andrew Ryan, Ankit Ramchandani, Annie Dong, Annie Franco, Anuj Goyal, Aparajita Saraf, Arkabandhu Chowdhury, Ashley Gabriel, Ashwin Bharambe, Assaf Eisenman, Azadeh
  Yazdan, Beau James, Ben Maurer, Benjamin Leonhardi, Bernie Huang, Beth Loyd, Beto~De Paola, Bhargavi Paranjape, Bing Liu, Bo~Wu, Boyu Ni, Braden Hancock, Bram Wasti, Brandon Spence, Brani Stojkovic, Brian Gamido, Britt Montalvo, Carl Parker, Carly Burton, Catalina Mejia, Ce~Liu, Changhan Wang, Changkyu Kim, Chao Zhou, Chester Hu, Ching-Hsiang Chu, Chris Cai, Chris Tindal, Christoph Feichtenhofer, Cynthia Gao, Damon Civin, Dana Beaty, Daniel Kreymer, Daniel Li, David Adkins, David Xu, Davide Testuggine, Delia David, Devi Parikh, Diana Liskovich, Didem Foss, Dingkang Wang, Duc Le, Dustin Holland, Edward Dowling, Eissa Jamil, Elaine Montgomery, Eleonora Presani, Emily Hahn, Emily Wood, Eric-Tuan Le, Erik Brinkman, Esteban Arcaute, Evan Dunbar, Evan Smothers, Fei Sun, Felix Kreuk, Feng Tian, Filippos Kokkinos, Firat Ozgenel, Francesco Caggioni, Frank Kanayet, Frank Seide, Gabriela~Medina Florez, Gabriella Schwarz, Gada Badeer, Georgia Swee, Gil Halpern, Grant Herman, Grigory Sizov, Guangyi, Zhang, Guna
  Lakshminarayanan, Hakan Inan, Hamid Shojanazeri, Han Zou, Hannah Wang, Hanwen Zha, Haroun Habeeb, Harrison Rudolph, Helen Suk, Henry Aspegren, Hunter Goldman, Hongyuan Zhan, Ibrahim Damlaj, Igor Molybog, Igor Tufanov, Ilias Leontiadis, Irina-Elena Veliche, Itai Gat, Jake Weissman, James Geboski, James Kohli, Janice Lam, Japhet Asher, Jean-Baptiste Gaya, Jeff Marcus, Jeff Tang, Jennifer Chan, Jenny Zhen, Jeremy Reizenstein, Jeremy Teboul, Jessica Zhong, Jian Jin, Jingyi Yang, Joe Cummings, Jon Carvill, Jon Shepard, Jonathan McPhie, Jonathan Torres, Josh Ginsburg, Junjie Wang, Kai Wu, Kam~Hou U, Karan Saxena, Kartikay Khandelwal, Katayoun Zand, Kathy Matosich, Kaushik Veeraraghavan, Kelly Michelena, Keqian Li, Kiran Jagadeesh, Kun Huang, Kunal Chawla, Kyle Huang, Lailin Chen, Lakshya Garg, Lavender A, Leandro Silva, Lee Bell, Lei Zhang, Liangpeng Guo, Licheng Yu, Liron Moshkovich, Luca Wehrstedt, Madian Khabsa, Manav Avalani, Manish Bhatt, Martynas Mankus, Matan Hasson, Matthew Lennie, Matthias Reso, Maxim
  Groshev, Maxim Naumov, Maya Lathi, Meghan Keneally, Miao Liu, Michael~L. Seltzer, Michal Valko, Michelle Restrepo, Mihir Patel, Mik Vyatskov, Mikayel Samvelyan, Mike Clark, Mike Macey, Mike Wang, Miquel~Jubert Hermoso, Mo~Metanat, Mohammad Rastegari, Munish Bansal, Nandhini Santhanam, Natascha Parks, Natasha White, Navyata Bawa, Nayan Singhal, Nick Egebo, Nicolas Usunier, Nikhil Mehta, Nikolay~Pavlovich Laptev, Ning Dong, Norman Cheng, Oleg Chernoguz, Olivia Hart, Omkar Salpekar, Ozlem Kalinli, Parkin Kent, Parth Parekh, Paul Saab, Pavan Balaji, Pedro Rittner, Philip Bontrager, Pierre Roux, Piotr Dollar, Polina Zvyagina, Prashant Ratanchandani, Pritish Yuvraj, Qian Liang, Rachad Alao, Rachel Rodriguez, Rafi Ayub, Raghotham Murthy, Raghu Nayani, Rahul Mitra, Rangaprabhu Parthasarathy, Raymond Li, Rebekkah Hogan, Robin Battey, Rocky Wang, Russ Howes, Ruty Rinott, Sachin Mehta, Sachin Siby, Sai~Jayesh Bondu, Samyak Datta, Sara Chugh, Sara Hunt, Sargun Dhillon, Sasha Sidorov, Satadru Pan, Saurabh Mahajan,
  Saurabh Verma, Seiji Yamamoto, Sharadh Ramaswamy, Shaun Lindsay, Shaun Lindsay, Sheng Feng, Shenghao Lin, Shengxin~Cindy Zha, Shishir Patil, Shiva Shankar, Shuqiang Zhang, Shuqiang Zhang, Sinong Wang, Sneha Agarwal, Soji Sajuyigbe, Soumith Chintala, Stephanie Max, Stephen Chen, Steve Kehoe, Steve Satterfield, Sudarshan Govindaprasad, Sumit Gupta, Summer Deng, Sungmin Cho, Sunny Virk, Suraj Subramanian, Sy~Choudhury, Sydney Goldman, Tal Remez, Tamar Glaser, Tamara Best, Thilo Koehler, Thomas Robinson, Tianhe Li, Tianjun Zhang, Tim Matthews, Timothy Chou, Tzook Shaked, Varun Vontimitta, Victoria Ajayi, Victoria Montanez, Vijai Mohan, Vinay~Satish Kumar, Vishal Mangla, Vlad Ionescu, Vlad Poenaru, Vlad~Tiberiu Mihailescu, Vladimir Ivanov, Wei Li, Wenchen Wang, Wenwen Jiang, Wes Bouaziz, Will Constable, Xiaocheng Tang, Xiaojian Wu, Xiaolan Wang, Xilun Wu, Xinbo Gao, Yaniv Kleinman, Yanjun Chen, Ye~Hu, Ye~Jia, Ye~Qi, Yenda Li, Yilin Zhang, Ying Zhang, Yossi Adi, Youngjin Nam, Yu, Wang, Yu~Zhao, Yuchen Hao, Yundi
  Qian, Yunlu Li, Yuzi He, Zach Rait, Zachary DeVito, Zef Rosnbrick, Zhaoduo Wen, Zhenyu Yang, Zhiwei Zhao, and Zhiyu Ma.
\newblock The llama 3 herd of models, 2024.
\newblock URL \url{https://arxiv.org/abs/2407.21783}.

\bibitem[Gridach et~al.(2025)Gridach, Nanavati, Mack, Abidine, and Mendes]{gridach2025agentic}
Mourad Gridach, Jay Nanavati, Christina Mack, Khaldoun Zine~El Abidine, and Lenon Mendes.
\newblock Agentic {AI} for scientific discovery: A survey of progress, challenges, and future directions.
\newblock In \emph{Towards Agentic AI for Science: Hypothesis Generation, Comprehension, Quantification, and Validation}, 2025.
\newblock URL \url{https://openreview.net/forum?id=TyCYakX9BD}.

\bibitem[Huang et~al.(2025)Huang, Shao, Hou, Yang, Wang, Sheng, Sun, and Wang]{huang2025pcg}
Lan Huang, Yangguang Shao, Wenju Hou, Hui Yang, Yan Wang, Nan Sheng, Yinglu Sun, and Yao Wang.
\newblock Pcg-cam: Enhanced class activation map using principal components of gradients and its applications in brain mri.
\newblock \emph{Information Sciences}, 708:\penalty0 122046, 2025.

\bibitem[Jiang et~al.(2024)Jiang, Kachinthaya, Petryk, and Gandelsman]{jiang2024interpreting}
Nick Jiang, Anish Kachinthaya, Suzie Petryk, and Yossi Gandelsman.
\newblock Interpreting and editing vision-language representations to mitigate hallucinations.
\newblock \emph{arXiv preprint arXiv:2410.02762}, 2024.

\bibitem[Kannan et~al.(2024)Kannan, Venkatesh, and Min]{KannanPurdue}
Shyam~Sundar Kannan, Vishnunandan L.~N. Venkatesh, and Byung-Cheol Min.
\newblock Smart-llm: Smart multi-agent robot task planning using large language models.
\newblock In \emph{2024 IEEE/RSJ International Conference on Intelligent Robots and Systems (IROS)}, pages 12140--12147, 2024.
\newblock \doi{10.1109/IROS58592.2024.10802322}.

\bibitem[Kim et~al.(2025)Kim, Choi, and Lee]{kim2025promptflowintegrityprevent}
Juhee Kim, Woohyuk Choi, and Byoungyoung Lee.
\newblock Prompt flow integrity to prevent privilege escalation in llm agents, 2025.
\newblock URL \url{https://arxiv.org/abs/2503.15547}.

\bibitem[Kram{\'a}r et~al.(2024)Kram{\'a}r, Lieberum, Shah, and Nanda]{kramar2024atp}
J{\'a}nos Kram{\'a}r, Tom Lieberum, Rohin Shah, and Neel Nanda.
\newblock Atp*: An efficient and scalable method for localizing llm behaviour to components.
\newblock \emph{arXiv preprint arXiv:2403.00745}, 2024.

\bibitem[Kramár et~al.(2024)Kramár, Lieberum, Shah, and Nanda]{kramár2024atpefficientscalablemethod}
János Kramár, Tom Lieberum, Rohin Shah, and Neel Nanda.
\newblock Atp*: An efficient and scalable method for localizing llm behaviour to components, 2024.
\newblock URL \url{https://arxiv.org/abs/2403.00745}.

\bibitem[Krishnan(2025)]{krishnan2025aiagentsevolutionarchitecture}
Naveen Krishnan.
\newblock Ai agents: Evolution, architecture, and real-world applications, 2025.
\newblock URL \url{https://arxiv.org/abs/2503.12687}.

\bibitem[Lu et~al.(2024)Lu, Lu, Lange, Foerster, Clune, and Ha]{lu2024aiscientistfullyautomatedSakana}
Chris Lu, Cong Lu, Robert~Tjarko Lange, Jakob Foerster, Jeff Clune, and David Ha.
\newblock The ai scientist: Towards fully automated open-ended scientific discovery, 2024.
\newblock URL \url{https://arxiv.org/abs/2408.06292}.

\bibitem[McInnes et~al.(2018)McInnes, Healy, and Melville]{mcinnes2018umap}
Leland McInnes, John Healy, and James Melville.
\newblock Umap: Uniform manifold approximation and projection for dimension reduction.
\newblock \emph{arXiv preprint arXiv:1802.03426}, 2018.

\bibitem[Meng et~al.(2022)Meng, Bau, Andonian, and Belinkov]{meng2022locating}
Kevin Meng, David Bau, Alex~J Andonian, and Yonatan Belinkov.
\newblock Locating and editing factual associations in {GPT}.
\newblock In Alice~H. Oh, Alekh Agarwal, Danielle Belgrave, and Kyunghyun Cho, editors, \emph{Advances in Neural Information Processing Systems}, 2022.
\newblock URL \url{https://openreview.net/forum?id=-h6WAS6eE4}.

\bibitem[Nelson~Elhage and Olah(2023)]{anthropicBasisExistence}
Robert~Lasenby Nelson~Elhage and Christopher Olah.
\newblock {Privileged Bases in the Transformer Residual Stream}.
\newblock \emph{Transformer Circuits Thread}, 2023.
\newblock URL \url{https://transformer-circuits.pub/2023/privileged-basis/index.html}.

\bibitem[Nishu et~al.(2025)Nishu, Mehta, Abnar, Farajtabar, Horton, Najibi, Nabi, Cho, and Naik]{nishu2025denseApple}
Kumari Nishu, Sachin Mehta, Samira Abnar, Mehrdad Farajtabar, Maxwell Horton, Mahyar Najibi, Moin Nabi, Minsik Cho, and Devang Naik.
\newblock From dense to dynamic: Token-difficulty driven moefication of pre-trained llms.
\newblock \emph{arXiv preprint arXiv:2502.12325}, 2025.

\bibitem[Olsson et~al.(2022)Olsson, Elhage, Nanda, Joseph, DasSarma, Henighan, Mann, Askell, Bai, Chen, et~al.]{olsson2022context}
Catherine Olsson, Nelson Elhage, Neel Nanda, Nicholas Joseph, Nova DasSarma, Tom Henighan, Ben Mann, Amanda Askell, Yuntao Bai, Anna Chen, et~al.
\newblock In-context learning and induction heads.
\newblock \emph{arXiv preprint arXiv:2209.11895}, 2022.

\bibitem[OpenAI(2021)]{openAI_codex}
OpenAI.
\newblock {OpenAI Codex}, 2021.
\newblock URL \url{https://github.com/openai/codex}.

\bibitem[Park et~al.(2025)Park, Du, Yeh, Wang, and Li]{park2025steer}
Seongheon Park, Xuefeng Du, Min-Hsuan Yeh, Haobo Wang, and Yixuan Li.
\newblock Steer llm latents for hallucination detection.
\newblock \emph{arXiv preprint arXiv:2503.01917}, 2025.

\bibitem[Qwen et~al.(2025)Qwen, :, Yang, Yang, Zhang, Hui, Zheng, Yu, Li, Liu, Huang, Wei, Lin, Yang, Tu, Zhang, Yang, Yang, Zhou, Lin, Dang, Lu, Bao, Yang, Yu, Li, Xue, Zhang, Zhu, Men, Lin, Li, Tang, Xia, Ren, Ren, Fan, Su, Zhang, Wan, Liu, Cui, Zhang, and Qiu]{qwen2025qwen25technicalreport}
Qwen, :, An~Yang, Baosong Yang, Beichen Zhang, Binyuan Hui, Bo~Zheng, Bowen Yu, Chengyuan Li, Dayiheng Liu, Fei Huang, Haoran Wei, Huan Lin, Jian Yang, Jianhong Tu, Jianwei Zhang, Jianxin Yang, Jiaxi Yang, Jingren Zhou, Junyang Lin, Kai Dang, Keming Lu, Keqin Bao, Kexin Yang, Le~Yu, Mei Li, Mingfeng Xue, Pei Zhang, Qin Zhu, Rui Men, Runji Lin, Tianhao Li, Tianyi Tang, Tingyu Xia, Xingzhang Ren, Xuancheng Ren, Yang Fan, Yang Su, Yichang Zhang, Yu~Wan, Yuqiong Liu, Zeyu Cui, Zhenru Zhang, and Zihan Qiu.
\newblock Qwen2.5 technical report, 2025.
\newblock URL \url{https://arxiv.org/abs/2412.15115}.

\bibitem[Rodrigues et~al.(2024)Rodrigues, Boutry, and Najman]{rodrigues2024unsupervised}
Caroline~Mazini Rodrigues, Nicolas Boutry, and Laurent Najman.
\newblock Unsupervised discovery of interpretable visual concepts.
\newblock \emph{Information Sciences}, 661:\penalty0 120159, 2024.

\bibitem[Sharma and Raman(2024)]{vanshRAG}
Vansh Sharma and Venkat Raman.
\newblock A reliable knowledge processing framework for combustion science using foundation models.
\newblock \emph{Energy and AI}, 16:\penalty0 100365, 2024.
\newblock ISSN 2666-5468.
\newblock \doi{https://doi.org/10.1016/j.egyai.2024.100365}.

\bibitem[Szanda{\l}a(2023)]{szandala2023unlocking}
Tomasz Szanda{\l}a.
\newblock Unlocking the black box of cnns: Visualising the decision-making process with prism.
\newblock \emph{Information Sciences}, 642:\penalty0 119162, 2023.

\bibitem[Templeton et~al.(2024)]{templeton2023_claude}
Adly Templeton et~al.
\newblock {Scaling Monosemanticity: Extracting Interpretable Features from Claude 3 Sonnet}.
\newblock \emph{Transformer Circuits Thread}, 2024.
\newblock URL \url{https://transformer-circuits.pub/2024/scaling-monosemanticity}.

\bibitem[Tian et~al.(2024)Tian, Gao, Zhang, Chen, Fan, Guo, Haas, Ji, Krongchon, Li, Liu, Luo, Ma, Tong, Trinh, Tian, Wang, Wu, Xiong, Yin, Zhu, Lieret, Lu, Liu, Du, Tao, Press, Callan, Huerta, and Peng]{tian2024scicoderesearchcodingbenchmark}
Minyang Tian, Luyu Gao, Shizhuo~Dylan Zhang, Xinan Chen, Cunwei Fan, Xuefei Guo, Roland Haas, Pan Ji, Kittithat Krongchon, Yao Li, Shengyan Liu, Di~Luo, Yutao Ma, Hao Tong, Kha Trinh, Chenyu Tian, Zihan Wang, Bohao Wu, Yanyu Xiong, Shengzhu Yin, Minhui Zhu, Kilian Lieret, Yanxin Lu, Genglin Liu, Yufeng Du, Tianhua Tao, Ofir Press, Jamie Callan, Eliu Huerta, and Hao Peng.
\newblock Scicode: A research coding benchmark curated by scientists, 2024.
\newblock URL \url{https://arxiv.org/abs/2407.13168}.

\bibitem[Vaswani et~al.(2017)Vaswani, Shazeer, Parmar, Uszkoreit, Jones, Gomez, Kaiser, and Polosukhin]{Vaswaniattentionisall}
Ashish Vaswani, Noam Shazeer, Niki Parmar, Jakob Uszkoreit, Llion Jones, Aidan~N Gomez, \L{}ukasz Kaiser, and Illia Polosukhin.
\newblock {Attention is All you Need}.
\newblock \emph{Advances in Neural Information Processing Systems}, 2017.

\bibitem[Vig et~al.(2020)Vig, Gehrmann, Belinkov, Qian, Nevo, Singer, and Shieber]{activationPatching}
Jesse Vig, Sebastian Gehrmann, Yonatan Belinkov, Sharon Qian, Daniel Nevo, Yaron Singer, and Stuart Shieber.
\newblock Investigating gender bias in language models using causal mediation analysis.
\newblock In H.~Larochelle, M.~Ranzato, R.~Hadsell, M.F. Balcan, and H.~Lin, editors, \emph{Advances in Neural Information Processing Systems}, volume~33, pages 12388--12401. Curran Associates, Inc., 2020.
\newblock URL \url{https://proceedings.neurips.cc/paper_files/paper/2020/file/92650b2e92217715fe312e6fa7b90d82-Paper.pdf}.

\bibitem[Wang et~al.(2023)Wang, Xie, Jiang, Mandlekar, Xiao, Zhu, Fan, and Anandkumar]{wang2023voyageropenendedembodiedagent}
Guanzhi Wang, Yuqi Xie, Yunfan Jiang, Ajay Mandlekar, Chaowei Xiao, Yuke Zhu, Linxi Fan, and Anima Anandkumar.
\newblock Voyager: An open-ended embodied agent with large language models, 2023.
\newblock URL \url{https://arxiv.org/abs/2305.16291}.

\bibitem[Wang et~al.(2022)Wang, Variengien, Conmy, Shlegeris, and Steinhardt]{wang2022interpretability}
Kevin Wang, Alexandre Variengien, Arthur Conmy, Buck Shlegeris, and Jacob Steinhardt.
\newblock Interpretability in the wild: a circuit for indirect object identification in gpt-2 small.
\newblock \emph{arXiv preprint arXiv:2211.00593}, 2022.

\bibitem[Wang et~al.(2024{\natexlab{a}})Wang, Ma, Feng, Zhang, Yang, Zhang, Chen, Tang, Chen, Lin, et~al.]{wang2024survey}
Lei Wang, Chen Ma, Xueyang Feng, Zeyu Zhang, Hao Yang, Jingsen Zhang, Zhiyuan Chen, Jiakai Tang, Xu~Chen, Yankai Lin, et~al.
\newblock A survey on large language model based autonomous agents.
\newblock \emph{Frontiers of Computer Science}, 18\penalty0 (6):\penalty0 186345, 2024{\natexlab{a}}.

\bibitem[Wang et~al.(2025)Wang, Jiao, Zhu, Chen, He, Chu, Gao, Wang, and Ma]{act}
Tianlong Wang, Xianfeng Jiao, Yinghao Zhu, Zhongzhi Chen, Yifan He, Xu~Chu, Junyi Gao, Yasha Wang, and Liantao Ma.
\newblock Adaptive activation steering: A tuning-free llm truthfulness improvement method for diverse hallucinations categories.
\newblock In \emph{Proceedings of the ACM on Web Conference 2025}, WWW '25, page 2562–2578, New York, NY, USA, 2025. Association for Computing Machinery.
\newblock ISBN 9798400712746.
\newblock \doi{10.1145/3696410.3714640}.
\newblock URL \url{https://doi.org/10.1145/3696410.3714640}.

\bibitem[Wang et~al.(2024{\natexlab{b}})Wang, Chen, Yuan, Zhang, Li, Peng, and Ji]{wang2024executableCoadeAct}
Xingyao Wang, Yangyi Chen, Lifan Yuan, Yizhe Zhang, Yunzhu Li, Hao Peng, and Heng Ji.
\newblock Executable code actions elicit better llm agents.
\newblock In \emph{ICML}, 2024{\natexlab{b}}.

\bibitem[Xiao et~al.(2023)Xiao, Tian, Chen, Han, and Lewis]{xiao2023efficient}
Guangxuan Xiao, Yuandong Tian, Beidi Chen, Song Han, and Mike Lewis.
\newblock Efficient streaming language models with attention sinks.
\newblock \emph{arXiv preprint arXiv:2309.17453}, 2023.

\bibitem[Yang et~al.(2024)Yang, Raman, Shah, and Tellex]{ZiyiBrownLLMRobots}
Ziyi Yang, Shreyas~S. Raman, Ankit Shah, and Stefanie Tellex.
\newblock Plug in the safety chip: Enforcing constraints for llm-driven robot agents.
\newblock In \emph{2024 IEEE International Conference on Robotics and Automation (ICRA)}, pages 14435--14442, 2024.
\newblock \doi{10.1109/ICRA57147.2024.10611447}.

\bibitem[Yeo et~al.(2025)Yeo, Mao, Abdar, Cambria, and Satapathy]{yeo2025debiasing}
Wei~Jie Yeo, Rui Mao, Moloud Abdar, Erik Cambria, and Ranjan Satapathy.
\newblock Debiasing clip: Interpreting and correcting bias in attention heads.
\newblock \emph{arXiv preprint arXiv:2505.17425}, 2025.

\bibitem[Yona et~al.(2025)Yona, Shumailov, Hayes, Barbero, and Gandelsman]{yona2025interpretingDeepMind}
Itay Yona, Ilia Shumailov, Jamie Hayes, Federico Barbero, and Yossi Gandelsman.
\newblock Interpreting the repeated token phenomenon in large language models.
\newblock \emph{arXiv preprint arXiv:2503.08908}, 2025.

\bibitem[Zhang and Nanda(2024)]{zhang2024towardsActivation}
Fred Zhang and Neel Nanda.
\newblock Towards best practices of activation patching in language models: Metrics and methods.
\newblock In \emph{The Twelfth International Conference on Learning Representations}, 2024.
\newblock URL \url{https://openreview.net/forum?id=Hf17y6u9BC}.

\bibitem[Zhang et~al.(2023)Zhang, Chen, Shen, Ding, Tenenbaum, and Gan]{zhang2023planning}
Shun Zhang, Zhenfang Chen, Yikang Shen, Mingyu Ding, Joshua~B. Tenenbaum, and Chuang Gan.
\newblock Planning with large language models for code generation.
\newblock In \emph{The Eleventh International Conference on Learning Representations}, 2023.
\newblock URL \url{https://openreview.net/forum?id=Lr8cOOtYbfL}.

\end{thebibliography}

\newpage
\appendix

\section*{Appendix}
\section{Results}

\subsection{Model Preference (Extended)} \label{appendix:resutls_models}
The following figure illustrates the code‐generation language distributions for Llama-3.3-70B-Instruct, Qwen2.5-Coder-32B-Instruct, and QwQ-32B. In each model, Julia and Python dominate the outputs, while Java and CPP appear only marginally.
\begin{figure}[!htb]
  \centering
  \includegraphics[width=1.0\linewidth]{ 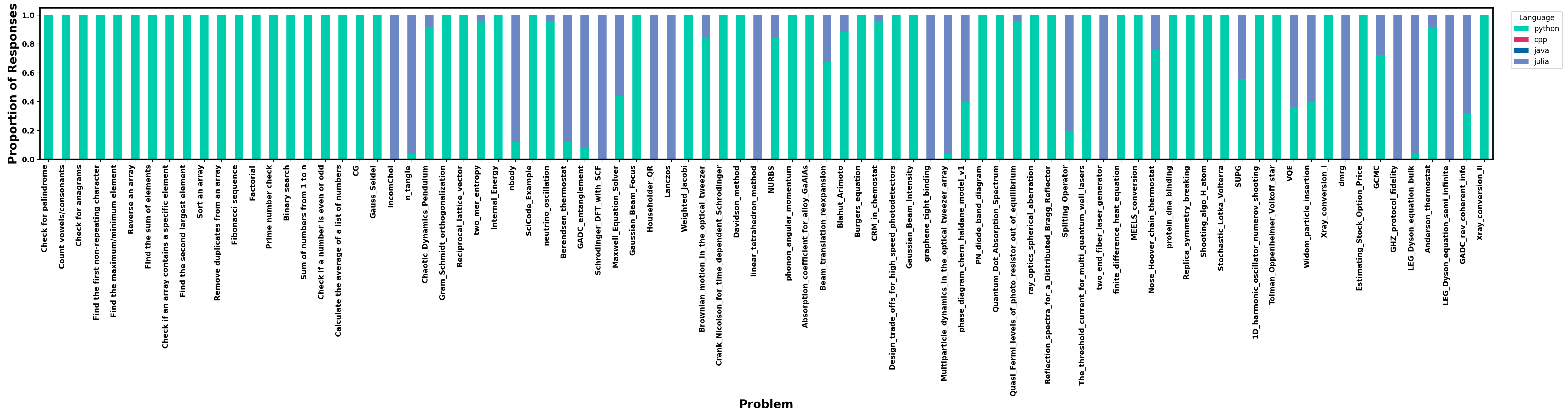}
  \caption{Programming‐language preferences of Llama-3.3-70B-Instruct on a set of coding tasks (25 runs per task at T = 1.0).}
  \label{fig:ll70b}
\end{figure}

\begin{figure}[!htb]
  \centering
  \includegraphics[width=1.0\linewidth]{ 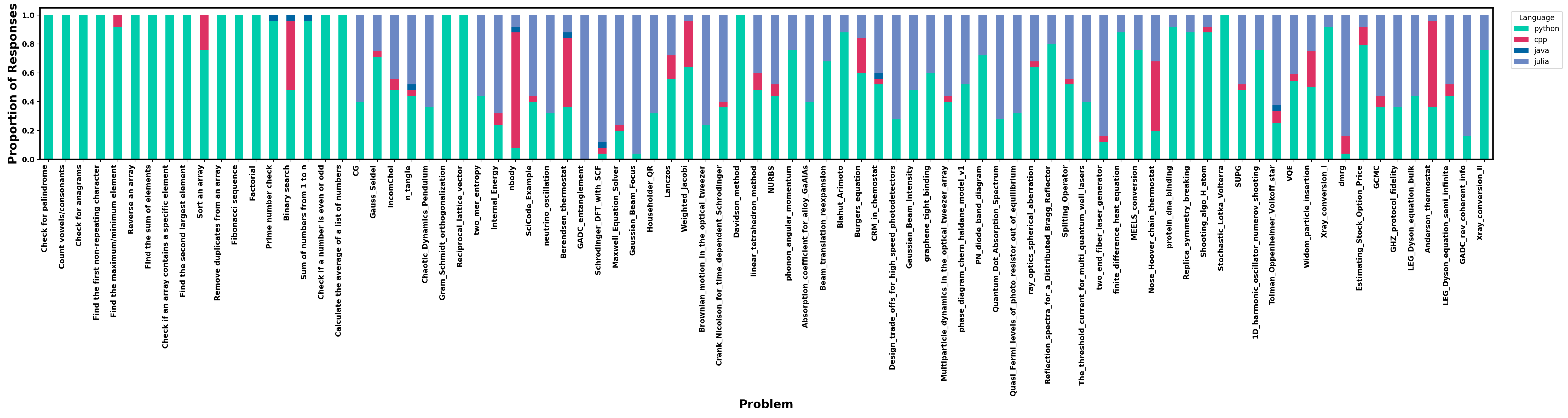}
  \caption{Programming‐language preferences of Qwen2.5-Coder-32B-Instruct on a set of coding tasks (25 runs per task at T = 1.0).}
  \label{fig:Q32b}
\end{figure}

\begin{figure}[!htb]
  \centering
  \includegraphics[width=1.0\linewidth]{ 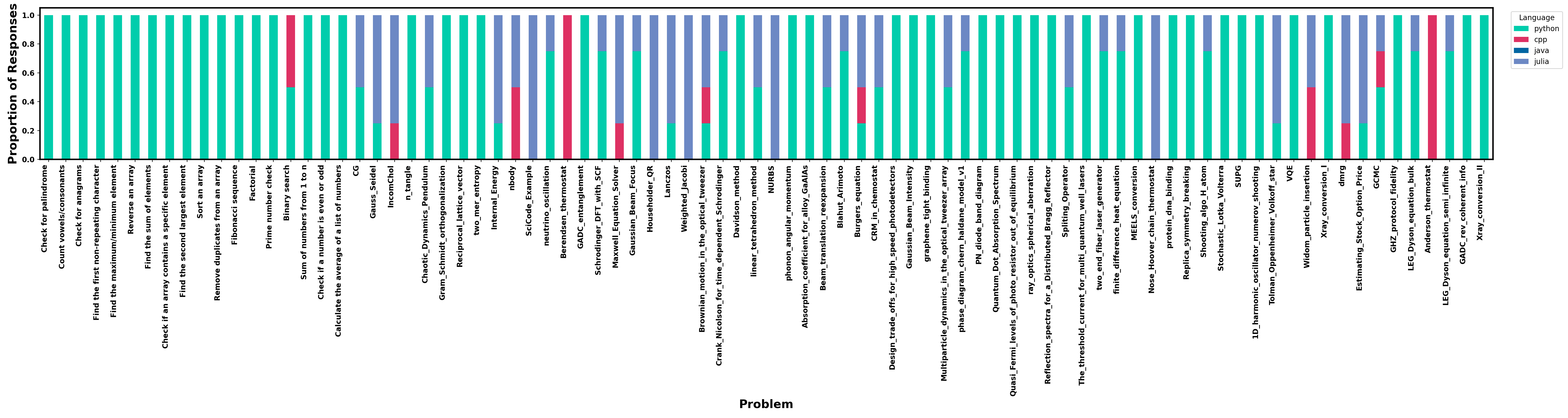}
  \caption{Programming‐language preferences of QwQ-32B on a set of coding tasks (25 runs per task at T = 1.0).}
  \label{fig:QwQ32b}
\end{figure}

\subsection{Activation Maps}

\begin{figure} [H]
  \centering
  \includegraphics[width=1.0\linewidth]{ 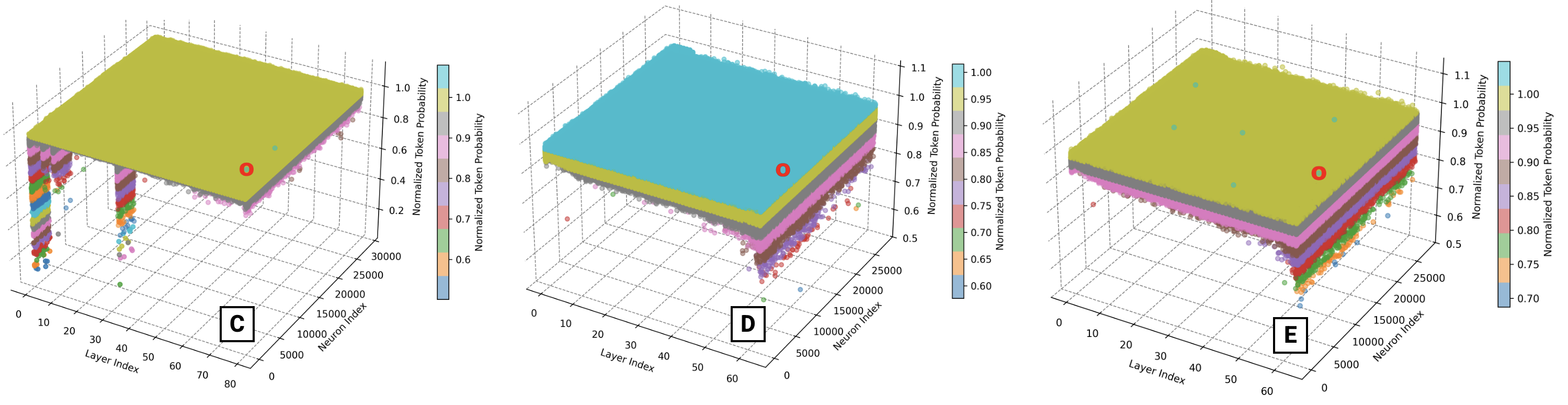}
  \caption{Activation maps for C. Llama-3.3-70B-Instruct, D. QwQ-32B and E. Qwen2.5-Coder-32B-Instruct models respectively. Neuron with highest activation probability is marked with a red circle for CPP token. }
  \label{fig:3dmaps2}
\end{figure}

\subsection{Steering Vectors for Llama-3.3-70B-Instruct Model}
\begin{figure} [H]
  \centering
  \includegraphics[width=1.0\linewidth]{ 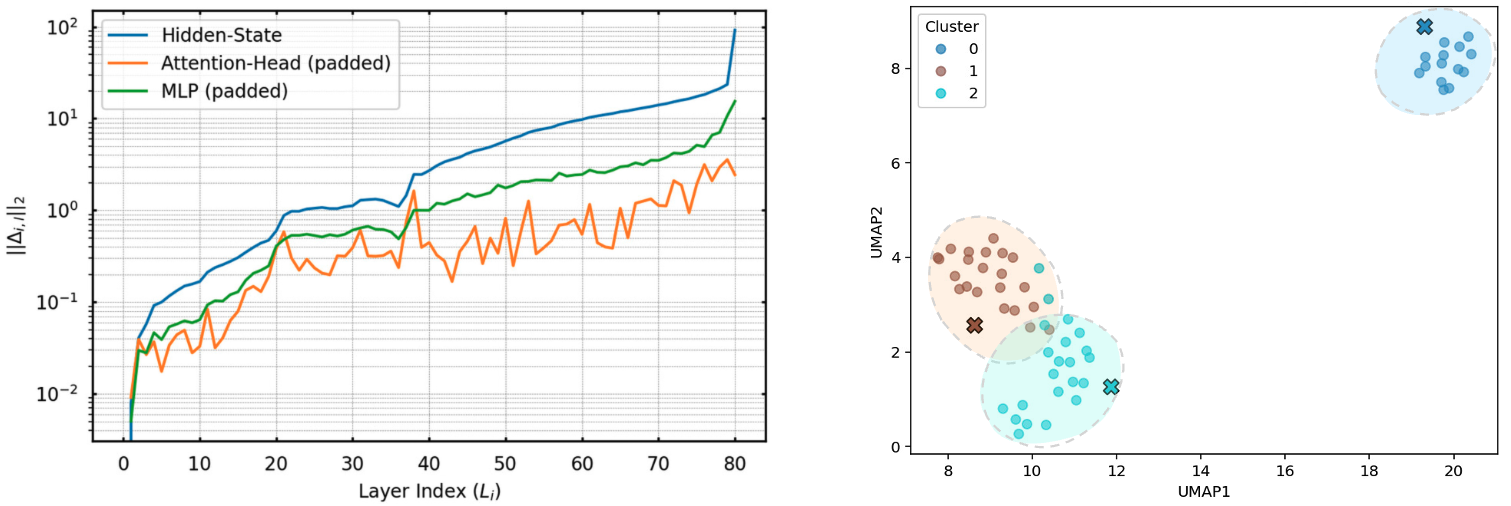}
  \caption{Results for Llama‑3.3‑70B‑Instruct model. Left: mean $\ell_2$‑norm of per‐layer CPP–Python difference vectors as a function of layer index. Right: UMAP projection of the flattened head‑difference vectors, colored by four K‑means clusters; crosses mark the cluster centroids.}
  \label{fig:diff-norm-clusters_70B}
\end{figure}

Figure \ref{fig:diff-norm-clusters_70B} illustrates two complementary views of the CPP – Python steering signal in the 70B LLaMA model. On the left, the mean norm of per-layer difference vectors grows steadily with depth, showing that later layers carry substantially stronger language-style cues (hidden-state \> MLP \> attention-head). On the right, a UMAP embedding of the flattened head-difference vectors reveals three well-separated clusters of prompts, each centroid (x) capturing a distinct style subgroup. Together, these plots confirm that (1) deeper layers encode more pronounced style differences and (2) prompt-level style variations can be compactly represented by a small number of steering directions.

\subsection{Code Generation Examples}

\begin{tcbraster}[raster equal height,raster valign=top, raster halign=center,raster columns=1, raster rows=1,
colframe=white,colback=white,colbacktitle=teal!50!black,] \label{block:photonCode}
\begin{tcolorbox}[title = Activated Neuron Llama‑3.2‑3B‑Instruct Model,colback=black!5!white,colframe=teal!75!black,fonttitle=\bfseries,] 
\tiny{ 
\begin{verbatim}
#include <iostream>
#include <vector>
#include <cmath>
#include <complex>
#include <queue>

void boson_distribution(double temperature, double* omega_r, double* omega_i, double* sum = nullptr, int N, 
int total_modes = -1, int N phonons = -1) {
    for (int nu = 0; nu < N; nu++) {
        for (int i = 0; i < N phonons; i++) {
            double numerator = std::exp(-omega_r[i] / (temperature * TMath_Pow(2.718281828459045, TMath_Exp(1.0)))+1.0);
            double denominator = 1;

            double exponent = 0.0;
            for (int q_index1 = 0; q_index1 < N phonons; q_index1++) {
                for (int n1 = 0; n1 <= 2; n1++) {
                    exponent += std::exp(-(omega_i[q_index1] * n1));
                }
            }
            denominator = std::sqrt(exponent);

            if (sum!= nullptr) {
                sum[i] = numerator / denominator;
            } else {
                std::cout << numerator / denominator << std::endl;
            }
        }
    }
}

void calculate_phonon_angular_momentum(const double *frequencies, const double polarization_vector[3][3], 
int index_x, int index_y, int index_z, double temperature) {
double N_Bohr_Q *= 0;
int N phonons = 3;
double n_boson = 100.0;

double omega_r[100] = {0};
double omega_i[10] = {0};

for (int i = 0; i < N phonons; ++i) {
    omega_r[i] = frequencies[i];
    for (int k = 0; k < 3; ++k) {
        for (int j = 0; j < 3; ++j) {
            omega_i[i] += (1 / n_boson) * std::sqrt(std::pow(polarization_vector[k][j][0] * frequencies[i] 
            + n_Bohr_Q * omega_r[i] / sqrt(1 + 1),3)+
                    std::pow(polarization_vector[k][j][1] * frequencies[i] 
                    + n_Bohr_Q * omega_r[i] / sqrt(1 +orus),3)
            + std::pow(polarization_vector[k][j][2] * frequencies[i] + n_Bohr_Q * omega_r[i] / sqrt(1 + r
                    std::sqrt(omega
                    i
                    ))));
        }
    }
}
double sum = 0;
for (int q_index1 = 0; q_index1 < 3; ++q_index1) {
    double temp_sumsum[q_index1] = 0;
    for (int nu = 0; nu < N phonons; ++nu) {
        temp_sumsum[q_index1] += omega_i[nu] + 0.5;
        temp_sumsum[q_index1] *= polarization_vector[q_index1][0][0];
        temp_sumsum[q_index1] *= polarization_vector[q_index1][1][1];
        temp_sumsum[q_index1] *= polarization_vector[q_index1][2][2];
    }
    sum += temp_sumsum[q_index1];
}
double M_alpha[3][3] = {{0, -1, 0},
                      {1, 0, 0},
                      {0, 0, -1}};
double unit_matrix[3][3] = {1,0,0,0,1,0,0,0,1,
                         0,1,0,0,0,1,0,0,1,
                         0,0,1,0,0,0,1,0,1};

for (int alpha = 0; alpha < 3; alpha++) {
    double unit_matrix_new[3][3] = {};
    for (int i = 0; i < 3; ++i) {
        for (int j = 0; j < 3; ++j) {
            unit_matrix_new[i][j] = unit_matrix[i][j] * M_alpha[alpha][i] * polarization_vector[alpha][j];
        }
    }
    if (alpha == index_x) sum /= 16;
    if (alpha == index_y) sum /= 16;
    if (alpha == index_z) sum /= 16;
}
double L_alpha = 1.054571817*sum;
printf("%.15lf", L_alpha);
}
\end{verbatim}
}
\end{tcolorbox}
\begin{tcolorbox}[boxsep=0pt,boxrule=0pt,colback=white,colframe=white,enhanced jigsaw,left=0mm,right=0mm,top=0mm,bottom=0mm]
\begin{tcolorbox}[title = Standard Llama‑3.2‑3B‑Instruct Model,colback=black!5!white,colframe=teal!75!black,fonttitle=\bfseries,] 
\tiny{
\begin{verbatim}
import numpy as np
import matplotlib.pyplot as plt
from scipy.constants import Boltzmann
from numpy.fft import fft, ifft

def bose_distribution(omega, beta, n=0):
    return (np.exp(beta * omega) - 1) / (np.exp(beta * omega) - 1) + n

def calculate_phonon_angular_momentum(phonon_frequencies, phonon_polarization_vectors, matrix_elements, temperature):
    M = (np.eye(len(phonon_frequencies)) * np.eye(3)
         + np.vandermonde([1, np.pi, np.pi/2],
                         np.vandermonde([2, np.pi/2, np.pi/2])))
    beta = 1 / (Boltzmann * temperature)
    Lx = np.sum(phonon_polarization_vectors[:, :, np.newaxis, np.newaxis] @
                M @ phonon_polarization_vectors
                [np.newaxis, :
                  , np.newaxis, np.newaxis] * bose_distribution(phonon_frequencies, beta, n=0) +
                (1/2) * np.eye(3) @ np.sum(
                    phonon_polarization_vectors[:, np.newaxis, np.newaxis, np.newaxis] *
                    phonon_polarization_vectors[np.newaxis, :, np.newaxis,
                                            :],
                    axis=2) * bose_distribution(phonon_frequencies, beta, n=0), axis=2))
    Ly = np.sum(phonon_polarization_vectors[:, np.newaxis, :
                      , np.newaxis] @ M @ phonon_polarization_vectors[:, ::-1, :, np.newaxis] 
                      * bose_distribution(phonon_frequencies, beta, n=0) +
                  (1/2) * np.sum(phonon_polarization_vectors[:, :, np.newaxis, np.newaxis] *
              phonon_polarization_vectors[np.newaxis, ::-1, :, np.newaxis] 
              * bose_distribution(phonon_frequencies, beta, n=0), axis=2))
    Lz = np.sum(phonon_polarization_vectors[:, np.newaxis, :
             , np.newaxis] @ np.swapaxes(
                 M, axis=0) @ phonon_polarization_vectors[:, ::-1, :, np.newaxis] \
        * bose_distribution(phonon_frequencies, beta, n=0) + (1/2) * np.sum(
            phonon_polarization_vectors[:, :, np.newaxis, np.newaxis] @
            phonon_polarization_vectors[np.newaxis, :, :, np.newaxis] *
            bose_distribution(phonon_frequencies, beta, n=0), axis=2))
    return Lx, Ly, Lz

# Test the function
omega = np.linspace(1, 10, 100)
photon polarization_vectors = np.random.rand(10, 3, 3)
np.savez(compress=True,"phonon_frequency.npy", phonon_frequencies=omega)
np.savez(
    compress=True, phonon_polarization_vectors=photon_polarization_vectors)
M = np.load("M.npy")
Lx, Ly, Lz = calculate_phonon_angular_momentum(
    np.load("phonon_frequency.npy"), np.load("phonon_polarization_vectors.npz")
)["phonon_frequencies"]. squeeze(), np.load("phonon_frequency.npy"), np.load("phonon_polarization_vectors.npz")

plt.plot(Lx, Ly)
plt.show()
\end{verbatim}
}

\end{tcolorbox}
\end{tcolorbox}
\end{tcbraster}

\subsection{Code Generation with G-ACT} \label{appendix:codeGACT}

G-ACT is used to activate the CPP code generation behavior in the Llama 3B and 70B models for the following prompts:
\textbf{P1:} \textit{Create a function to solve the matrix equation $Ax=b$ using the Gauss-Seidel iteration. The function takes a matrix $A$ and a vector $b$ as inputs. The method involves splitting the matrix $A$ into the difference of two matrices, $A=M-N$. For Gauss-Seidel, $M=D-L$, where $D$ is the diagonal component of $A$ and $L$ is the lower triangular component of $A$. The function should implement the corresponding iterative solvers until the norm of the increment is less than the given tolerance, $||x_k - x_{k-1}||_{l_2}<\epsilon$.} \\
And \textbf{P2:} \textit{Write me a CFD code for the compressible Euler system of equations for a flow over a wedge at 30 degrees. Simply provide the code only. Assume air as the fluid at mach 3 inlet velocity and isentropic relations are used.}

As shown in the following code blocks, the models autonomously generate CPP code directly from the prompt, without any additional user guidance. Across repeated trials, it produces only CPP implementations, underscoring the robustness and effectiveness of the proposed framework.

\begin{tcbraster}[raster equal height,raster valign=top, raster halign=center,raster columns=1, raster rows=1,
colframe=white,colback=white,colbacktitle=teal!50!black,] \label{block:gact_p1}
\begin{tcolorbox}[title = G-ACT + Llama‑3.2‑3B‑Instruct Model for P1,colback=black!5!white,colframe=teal!75!black,fonttitle=\bfseries,breakable] 
\tiny{ 
\begin{verbatim}
#include <iostream>
#include <vector>

// Function to calculate the L2 norm (Euclidean norm)
double l2_norm(const std::vector<double>& vec) {
    double sum = 0;
    for (const auto& val : vec) { // const auto& allows us to avoid copying elements from'vec'
        sum += val * val;
    }
    return sqrt(sum);
}

// Function template to perform Gauss-Seidal iterations on Ax=b with A split into M=N=A-D
template<typename T> void gauss_seidel(T* A, int n, T* b, T tol=1e-10, int max_iter=10000) {

    if(n==1){
        T x[1]={};
        x[0]=b[0]/A[0];
        std::cout << "Solution: ";
        for(int i=0;i<n;++i){std::cout<<x[i]<<" ";}std::endl;
        return ;
    }

    T* x=new T[n]; // Allocate memory for solution vector

    bool converged=true;

    for(int iter=0;iter<max_iter&&converged;iter++){

        for (int j=n-1;j>=0;j--) {

            T sum=0;

            for ( int k=j+1;k<n;k++ ) {
                sum+=A[j*n+k]*x[k];
            }

            if(j!=n-1){
                sum-=A[j*(j+1)]*x[j];
            }

            x[j]=(b[j]-sum)/A[j*j];

        }

        T inc_l2=l2_norm(x);

        if(inc_l2<tol){
            converged=false;
        }

    }

     delete[] x; // Deallocate memory after use

}
\end{verbatim}
}
\end{tcolorbox}
\end{tcbraster}

\begin{tcbraster}[raster equal height,raster valign=top, raster halign=center,raster columns=1, raster rows=1,
colframe=white,colback=white,colbacktitle=teal!50!black,] \label{block:gact_p2}
\begin{tcolorbox}[title = G-ACT + Llama‑3.3‑70B‑Instruct Model for P2,colback=black!5!white,colframe=teal!75!black,fonttitle=\bfseries,breakable] 
\tiny{ 
\begin{verbatim}
#include <iostream>
#include <fstream>
#include <vector>
#include <cmath>
#include <iomanip>

using namespace std;

// Define constants
const double PI = 3.14159265358979323846;
const double GAMMA = 1.4; // Adiabatic index for air
const double MACH = 3.0; // Mach number
const double THETA = 30.0 * PI / 180.0; // Angle of wedge in radians
const double R = 287.0; // Gas constant for air
const double CP = 1005.0; // Specific heat capacity at constant pressure for air
const double CV = CP - R; // Specific heat capacity at constant volume for air

// Define grid parameters
const int NX = 100; // Number of grid points in x-direction
const int NY = 100; // Number of grid points in y-direction
const double LX = 10.0; // Length of domain in x-direction
const double LY = 10.0; // Length of domain in y-direction
const double DX = LX / (NX - 1); // Grid spacing in x-direction
const double DY = LY / (NY - 1); // Grid spacing in y-direction

// Define flow parameters
const double P0 = 101325.0; // Stagnation pressure
const double T0 = 288.0; // Stagnation temperature
const double U0 = MACH * sqrt(GAMMA * R * T0); // Inlet velocity
const double P1 = P0 * pow(1.0 + (GAMMA - 1.0) / 2.0 * MACH * MACH, -GAMMA / (GAMMA - 1.0)); // Inlet pressure
const double T1 = T0 / (1.0 + (GAMMA - 1.0) / 2.0 * MACH * MACH); // Inlet temperature
const double RHO1 = P1 / (R * T1); // Inlet density

// Define time parameters
const double DT = 0.001; // Time step
const double T_END = 10.0; // End time

// Define variables
double rho[NX][NY]; // Density
double u[NX][NY]; // x-velocity
double v[NX][NY]; // y-velocity
double p[NX][NY]; // Pressure
double e[NX][NY]; // Total energy

// Function to calculate fluxes
void calculateFluxes(double &Fx, double &Fy, double rho, double u, double v, double p, double e) {
    Fx = rho * u;
    Fy = rho * v;
}

// Function to calculate source terms
void calculateSourceTerms(double &Sx, double &Sy, double rho, double u, double v, double p, double e) {
    Sx = 0.0;
    Sy = 0.0;
}

// Function to update variables
void updateVariables(double &rho, double &u, double &v, double &p, 
double &e, double Fx, double Fy, double Sx, double Sy) {
    rho += DT * (Sx - (Fx / DX + Fy / DY));
    u += DT * (Sx / rho - (Fx * u / (rho * DX) + Fy * u / (rho * DY)));
    v += DT * (Sy / rho - (Fx * v / (rho * DX) + Fy * v / (rho * DY)));
    p += DT * (Sx * u - (Fx * p / (rho * DX) + Fy * p / (rho * DY)));
    e += DT * (Sx * u - (Fx * e / (rho * DX) + Fy * e / (rho * DY)));
}

int main() {
    // Initialize variables
    for (int i = 0; i < NX; i++) {
        for (int j = 0; j < NY; j++) {
            rho[i][j] = RHO1;
            u[i][j] = U0;
            v[i][j] = 0.0;
            p[i][j] = P1;
            e[i][j] = P1 / (GAMMA - 1.0) + 0.5 * RHO1 * (U0 * U0 + 0.0 * 0.0);
        }
    }

    // Time loop
    for (double t = 0.0; t < T_END; t += DT) {
        // Calculate fluxes and source terms
        for (int i = 0; i < NX; i++) {
            for (int j = 0; j < NY; j++) {
                double Fx, Fy, Sx, Sy;
                calculateFluxes(Fx, Fy, rho[i][j], u[i][j], v[i][j], p[i][j], e[i][j]);
                calculateSourceTerms(Sx, Sy, rho[i][j], u[i][j], v[i][j], p[i][j], e[i][j]);

                // Update variables
                updateVariables(rho[i][j], u[i][j], v[i][j], p[i][j], e[i][j], Fx, Fy, Sx, Sy);
            }
        }

        // Apply boundary conditions
        for (int j = 0; j < NY; j++) {
            rho[0][j] = RHO1;
            u[0][j] = U0;
            v[0][j] = 0.0;
            p[0][j] = P1;
            e[0][j] = P1 / (GAMMA - 1.0) + 0.5 * RHO1 * (U0 * U0 + 0.0 * 0.0);
        }

        for (int i = 0; i < NX; i++) {
            rho[i][0] = RHO1;
            u[i][0] = U0 * sin(THETA);
            v[i][0] = U0 * cos(THETA);
            p[i][0] = P1;
            e[i][0] = P1 / (GAMMA - 1.0) + 0.5 * RHO1 * (U0 * U0 + 0.0 * 0.0);
        }

        // Output results
        if (t % 1.0 == 0.0) {
            cout << "Time: " << t << endl;
            for (int i = 0; i < NX; i++) {
                for (int j = 0; j < NY; j++) {
                    cout << rho[i][j] << " " << u[i][j] << " " << v[i][j] 
                    << " " << p[i][j] << " " << e[i][j] << endl;
                }
            }
        }
    }

    return 0;
}
\end{verbatim}
}
\end{tcolorbox}
\end{tcbraster}

\section{Prompts} \label{sec:prompts}
Following prompts were used for different tasks conducted in this study.

\begin{tcbraster}[raster valign=top, raster halign=center,raster columns=1, raster rows=1, colframe=white,colback=white,colbacktitle=black!50!white,] \label{block:prompts}

\begin{tcolorbox}[title = Prompt for Language Preference Task ,colback=black!5!white,colframe=gray!75!black,fonttitle=\bfseries, width = 0.475\linewidt]
\footnotesize{You are given a coding problem consisting of a problem name and a main description.
    Your task is to select one programming language from the following list that you think is best suited to solve the problem:
    ['python', 'cpp', 'julia', 'java']
    Understand the scale of the problem and the need, the problem could be scientific or general. You will be provided with the problem description. 
    
    Important:
    - Your response should consist of only the name of the chosen language.
    - Do not include any additional text or explanation.
    
    Example:
    "problem name": "Reverse a string",
    "problem description main": "Write a program to reverse a given string (e.g., 'hello' becomes 'olleh')."
    Expected Answer (for example):
    one programming language
    
    Given Problem:
    "problem name": "\{problem name\}",
    "problem description main": "\{problem description main\}"
    
    Expected one word Answer: 
}
\end{tcolorbox}

\begin{tcolorbox}[title = Prompt for Activation Testing: Language Preference Task, colback=black!5!white,colframe=gray!75!black, fonttitle=\bfseries, width = 0.475\linewidth]
\footnotesize{You are given a coding problem consisting of a problem name and a main description.
    Your task is to select one programming language from the following list and provide the code for the problem:
    ['python', 'cpp', 'julia', 'java']
    Understand the scale of the problem and the need, the problem could be scientific or general. You will be provided with the problem description. 
    
    Important:
    - Your response should consist of the code only.
    - Do not include any additional text or explanation.
    
    Given Problem:
    "problem name": "\{problem name\}",
    "problem description main": "\{problem description main\}"
    Expected Answer: }
\end{tcolorbox}

\begin{tcolorbox}[title = Prompt for Activation Testing: Code Generation  Task, colback=black!5!white,colframe=gray!75!black, fonttitle=\bfseries, width = 0.475\linewidth]
\footnotesize{You are given a coding problem consisting of a problem name and a main description.
    Your task is to select one programming language from the following list and provide the code for the problem:
    ['python', 'cpp']
    Understand the scale of the problem and the need, the problem could be scientific or general. You will be provided with the problem description.
    Important:
    - Your response should consist of the code only.
    - Do not include any additional text or explanation.
    
    Given Problem:
    "problem name": "\{problem name\}",
    "problem description main": "\{problem description main\}"
    Expected Answer: }
\end{tcolorbox}

\end{tcbraster}



\end{document}